\documentclass[journal]{IEEEtran}
\usepackage{amsmath,amsfonts}
\usepackage{algorithmic}
\usepackage{algorithm}
\usepackage{array}
\usepackage[caption=false,font=normalsize,labelfont=sf,textfont=sf]{subfig}
\usepackage{textcomp}
\usepackage{stfloats}
\usepackage{url}
\usepackage{verbatim}
\usepackage{graphicx}
\usepackage{cite}
\usepackage{amssymb}
\usepackage{threeparttable}
\usepackage{booktabs}
\usepackage{siunitx}
\usepackage{multirow}
\usepackage{hyperref}
\usepackage{xurl}
\usepackage{enumitem}
\usepackage{xcolor}
\usepackage{tikz}

\hypersetup{
    colorlinks=true,         
    linkcolor=black,         
    citecolor=black,         
    urlcolor=blue,           
    bookmarksnumbered=true,
    pdfpagemode=UseOutlines
}

\usepackage{pgfplots}
\usepackage{pgfplotstable} 
\usetikzlibrary{matrix}
\usepgfplotslibrary{groupplots}
\usetikzlibrary{shadows}
\pgfplotsset{compat=newest}

\usepgfplotslibrary{external}
\pgfplotsset{
    IEEEstyle/.style={
        width=\columnwidth,       
        height=6cm,               
        grid=major,               
        grid style={dashed, gray!30},
        tick label style={font=\scriptsize},
        label style={font=\scriptsize},
        title style={font=\scriptsize},
    }
}

\pgfplotscreateplotcyclelist{graph_precision_recall}{
    {blue, only marks, mark=+, mark size=2pt, thick},           
    {blue, no marks, solid, thick},                             
    {blue, no marks, dashed, thick},                            
    {red!70!black, no marks, solid, thick},                     
    {red!70!black, only marks, mark=x, mark size=2pt, thick},   
    {red!70!black, dashed, thick},                              
    {black!60, no marks, solid, thick},                         
    {black!60, no marks, dashed, thick},                        
    {black!60, only marks, mark=triangle, mark size=2.5pt, thick}, 
    {black, no marks, solid, ultra thick}                        
}

\pgfplotscreateplotcyclelist{graph_precision_recall_2}{
    {red!70!black, no marks, solid, thick},                     
    {red!70!black, only marks, mark=x, mark size=2pt, thick},   
    {red!70!black, dashed, thick},                              
    {black!60, no marks, solid, thick},                         
    {black!60, no marks, dashed, thick},                        
    {black!60, only marks, mark=triangle, mark size=2.5pt, thick}, 
    {black, no marks, solid, ultra thick}                        
}

\usepackage{xspace}
\newcommand{\etal}{\textit{et al.}\xspace}

\newenvironment{notablediff}{\begin{table}}{\end{table}}
\newenvironment{notablediff*}{\begin{table*}}{\end{table*}}
\newenvironment{nofigurediff*}{\begin{figure*}}{\end{figure*}}

\newcommand{\bullseye}{%
  \mbox{%
    \begin{tikzpicture}[baseline=-0.6ex, scale=1.0]
      \draw[line width=0.8pt] (0,0) circle (0.7ex);
      \draw[line width=0.8pt] (0,0) circle (0.3ex);
    \end{tikzpicture}%
  }%
  \xspace 
}

\begin{document}
\title{LEADER: Lightweight End-to-End Attention-Gated\\ Dual Encoder-Decoder Network for Robust Minutiae Extraction}
\author{Raffaele Cappelli and Matteo Ferrara
\thanks{The authors are with the Department of Computer Science and Engineering of the University of Bologna, Italy (e-mail: raffaele.cappelli@unibo.it, matteo.ferrara@unibo.it).}
\thanks{The \texttt{pyfing} package, including source code and pre-trained weights for LEADER, is available at \url{https://github.com/raffaele-cappelli/pyfing}.}
}

\markboth{ }%
{...}


\maketitle

\begin{abstract}
This paper introduces LEADER (Lightweight End-to-end Attention-gated Dual Encoder-decodeR), a neural network that maps raw fingerprint images to minutiae descriptors, including location, direction, and type. The proposed architecture integrates non-maximum suppression and angular decoding to enable end-to-end inference using only 0.9\,M parameters. 
It employs a novel ``Castle-Moat-Rampart'' ground-truth encoding and a dual encoder-decoder architecture, interconnected through an attention-gating mechanism. Experimental evaluations demonstrate state-of-the-art accuracy on plain fingerprints and robust cross-domain generalization to latent impressions across two benchmark datasets. Specifically, LEADER outperforms top-tier commercial systems, achieving $\mathbf{F_1}$-score gains of up to 34\% and 39\% over the closest competitor on NIST SD27 and SD302, respectively. Qualitative visual analysis suggests that the internal representations learned by the model align with established fingerprint domain concepts, such as segmentation masks, orientation fields, frequency maps, and skeletons. Inference requires only 15\,ms on GPU (average latency per image at batch size 32) and 322\,ms on CPU, outperforming leading commercial software in computational efficiency. The source code and pre-trained weights are publicly released to facilitate reproducibility.
\end{abstract}


\section{Introduction}
\label{sec:intro}
\begin{figure*}[t]
    \centering
    \includegraphics[width=\textwidth]{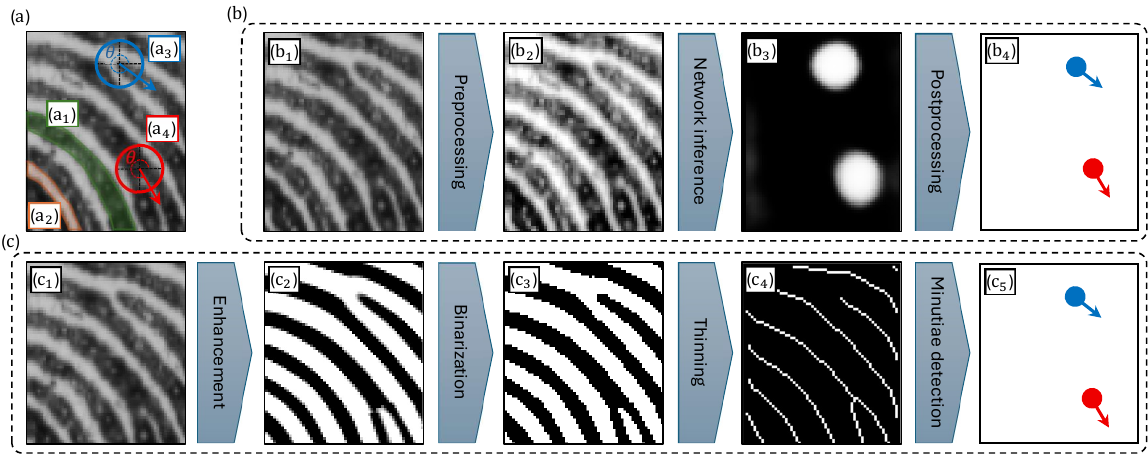}
    \caption{\textbf{Overview of fingerprint patterns and extraction paradigms.} (a) Local ridge structure highlighting a ridge ($\mathrm{a}_1$), a valley ($\mathrm{a}_2$), a ridge ending ($\mathrm{a}_3$), and a bifurcation ($\mathrm{a}_4$), with their corresponding directions $\theta$. (b) Conventional deep learning-based pipeline: the input image ($\mathrm{b}_1$) is typically preprocessed ($\mathrm{b}_2$) and fed into a neural network to generate heatmaps ($\mathrm{b}_3$), requiring off-graph postprocessing to obtain the final minutiae ($\mathrm{b}_4$). (c) Traditional extraction workflow: the input image ($\mathrm{c}_1$) is enhanced ($\mathrm{c}_2$), binarized ($\mathrm{c}_3$), and thinned to a ridge skeleton ($\mathrm{c}_4$) before connectivity analysis yields the minutiae ($\mathrm{c}_5$).}
    \label{fig:intro}
\end{figure*}

\IEEEPARstart{B}{iometric} recognition is a cornerstone of modern identity management, with fingerprint identification remaining one of the most widely adopted modalities due to its permanence, uniqueness, and high discriminative power~\cite{Maltoni2022}. At the core of automated fingerprint identification systems (AFIS) lies minutiae extraction, specifically the detection of ridge endings and bifurcations (Fig.~\ref{fig:intro}a). The spatial distribution, type, and direction of these local features form the basis for high-security fingerprint comparison. However, robust and precise minutiae extraction remains challenging, particularly for latent impressions or low-quality images~\cite{Maltoni2017}.

Historically, minutiae extraction relied on multi-stage pipelines involving image enhancement, binarization, and morphological thinning to produce a one-pixel-wide skeleton map~\cite{Maltoni2022}. Although effective on high-quality fingerprints, these traditional methods are prone to false detections and omissions when ridge patterns are degraded. To address these limitations, the field has shifted toward deep learning-based approaches. Early neural methods used sliding-window classifiers or patch-based analysis, which are computationally expensive and often lack global context. More recently, fully convolutional architectures have enabled the processing of entire images in a single forward pass.

Despite these advances, state-of-the-art (SOTA) models still require external preprocessing or significant post-inference operations, as summarized in Table~\ref{tabMethods}. Moreover, many recent architectures require large parameter counts to maintain accuracy, limiting their suitability for resource-constrained devices.

In this paper, we propose LEADER (Lightweight End-to-end Attention-gated Dual Encoder-decodeR), a fully convolutional model designed to map raw fingerprint images to comprehensive minutiae attributes covering location, direction, type, and quality score. This approach eliminates complex off-graph preprocessing and postprocessing steps, while delivering top-tier accuracy with remarkable parameter efficiency.

The main contributions of this work are as follows.
\begin{enumerate}  
    \item An \textit{end-to-end pipeline} that integrates the entire extraction process---including non-maximum suppression (NMS)---into the network, enabling direct readout of complete minutiae attributes.
    \item An \textit{adaptive ground-truth encoding}, termed Castle-Moat-Rampart (CMR), which uses spatially modulated weighting to disambiguate adjacent minutiae and improve localization robustness against labeling jitter.  
    \item A \textit{lightweight model} with two encoder-decoder components and a specialized attention gate, achieving high-accuracy minutiae extraction with only 0.9\,M parameters.
    \item A \textit{modernized convolutional backbone} incorporating separable convolutions, inverse bottlenecks, dilated kernels, and mixed pooling, optimized for accuracy and efficiency~\cite{Liu2022}.
    \item \textit{Cross-domain generalization to latent fingerprints} despite training exclusively on non-latent samples, with performance exceeding existing SOTA methods and commercial off-the-shelf (COTS) systems.
    \item A \textit{qualitative feature analysis}, showing that LEADER learns representations visually aligned with traditional fingerprint concepts---such as segmentation masks, orientation fields, frequency maps, and ridge and valley skeletons---without explicit supervision.
    \item An \textit{integrated open-source package} (\texttt{pyfing}), including pre-trained weights for immediate reproducibility and further research.
\end{enumerate}

The remainder of this paper is organized as follows. Sec.~\ref{sec:related} reviews the evolution of minutiae extraction methods. Sec.~\ref{sec:method} describes the proposed LEADER architecture, detailing its cascaded feature extraction, on-graph postprocessing, ground-truth encoding, and multi-task optimization. Sec.~\ref{sec:experiments} presents experimental evaluations, ablation studies, and qualitative feature analyses. Sec.~\ref{sec:conclusion} summarizes the contributions and outlines future directions. Supplementary Material (Appendices~\ref{sec:supp_architecture}--\ref{sec:supp_results}) provides architectural specifications, implementation details, training dynamics, boundary-filtering sensitivity analysis and extended visual comparisons of extraction accuracy.

\section{Related Works}
\label{sec:related}
Minutiae extraction has evolved from handcrafted morphological pipelines toward sophisticated deep learning architectures. This section reviews the milestones of this progression, from foundational methods to the latest neural approaches.

\subsection{Traditional Methods}
Earlier solutions are generally categorized into binarization-based frameworks and direct gray-scale approaches.

\subsubsection{Binarization-based Extraction}
These classic frameworks rely on a multi-stage sequential process (Fig.~\ref{fig:intro}c). The workflow begins with image enhancement and ridge-valley binarization, followed by morphological thinning to generate a one-pixel-wide skeleton~\cite{Maltoni2022}. Minutiae are then detected by analyzing local pixel connectivity, typically through the \textit{crossing number} technique, which identifies ridge endings and bifurcations by counting ridge-to-valley transitions in a $3 \times 3$ neighborhood~\cite{Maltoni2022}. Although effective for high‑quality fingerprints, this pipeline is notably vulnerable to noise. To mitigate artifacts such as ``hairy'' skeletons or spurs, regularization via morphological operators or adaptive filtering is often employed. However, these measures frequently fail to eliminate spurious minutiae, necessitating additional heuristic postprocessing that increases computational complexity.

\subsubsection{Direct Gray-scale Extraction}
Several methods operate directly on gray-scale intensities to mitigate the information loss and artifacts typical of binarization and thinning. These approaches generally follow one of two strategies. 
The first is \textit{ridge following}, exemplified by the algorithm in~\cite{Maio1997}, which tracks ridges along the orientation field to detect intensity maxima on orthogonal cross-sections. While this avoids binarization errors, the sequential nature of the process requires precise tuning and lacks robustness in low-quality areas. The second strategy, \textit{singularity modeling}, treats minutiae as local discontinuities in the ridge flow. Techniques based on parabolic symmetries~\cite{Fronthaler2008} or frequency modulation~\cite{Larkin2007} provide rigorous frameworks but often exhibit lower precision than modern learning-based alternatives.

\subsection{Deep Learning-based Methods}
Recent advancements have established deep learning as the dominant paradigm. Table~\ref{tabMethods} provides a taxonomy of representative methods, tracing the evolution from local classifiers to integrated pipelines.

\subsubsection{Patch-based Classifiers}
Early neural approaches~\cite{Sankaran2014, Jiang2016, Darlow2017} employ sliding-window strategies to classify individual patches as minutia or non-minutia. This binary output space precludes the inference of direction and type, while the overlapping scans result in high computational redundancy and lack of global context.

\subsubsection{Two-stage and Multi-pass Frameworks}
To incorporate broader context while maintaining precision, certain architectures adopt a multi-stage design. Methods in~\cite{Tang2017a, Nguyen2018a, Zhou2020} utilize an initial convolutional neural network (CNN) to identify candidate regions, followed by refinement (e.g., ROI-pooling or patch analysis) to regress coordinates and directions. Other approaches, such as~\cite{Cao2020}, perform multi-pass inference on externally enhanced inputs. Although more accurate than patch-based classifiers, these frameworks are computationally intensive due to their iterative execution and high parameter counts.

\subsubsection{Single-pass Pipelines}
The current state of the art involves CNNs designed for single-pass inference~\cite{Tang2017, Nguyen2020, Zhang2021, Feng2023}. These models process the entire fingerprint in a single forward pass (Fig.~\ref{fig:intro}b) to produce feature maps from which minutiae are derived. FingerNet~\cite{Tang2017} pioneered this shift by embedding traditional processing steps within differentiable layers. Despite their improved efficiency, these methods lack a unified inference-end-to-end design due to off-graph dependencies (e.g., external NMS) and typically provide incomplete attribute extraction by omitting minutiae type. The proposed LEADER architecture addresses these gaps by embedding all processing steps within a single on-graph computational flow, enabling the joint extraction of position, direction, and type. While backpropagation optimizes all learnable weights up to the task-specific feature maps, the integration of an on-graph NMS layer provides a seamless end-to-end pipeline. In this framework, the transition from dense feature maps to the final sparse minutiae list requires only a standard, non-heuristic thresholding readout.

\begin{table*}[t]
\centering
\begin{threeparttable}
\caption{Evolution of deep learning-based minutiae extraction methods and comparison with LEADER.}
\label{tabMethods}
\begin{tabular*}{\textwidth}{@{\extracolsep{\fill}} l c c c c S[table-format=3.1] c @{}}
\toprule
\textbf{Method} & \textbf{Single-pass} & \textbf{End-to-end pipeline} & \textbf{Minutiae direction} & \textbf{Minutiae type} & \textbf{Parameters (M)} & \textbf{Open-source}\\
\midrule
Sankaran \etal (2014)~\cite{Sankaran2014} & No\rlap{\tnote{1}} & No & No & No & 29.7 & No\\
Jiang \etal (2016)~\cite{Jiang2016} & No\rlap{\tnote{1}} & No & No & No & 11.2 & No\\
Darlow and Rosman (2017)~\cite{Darlow2017} & No\rlap{\tnote{1}} & No & No & No & 3.3 & No\\
Tang, Gao and Feng (2017)~\cite{Tang2017a} & No\rlap{\tnote{2}} & No & Yes & No & 108.3 & No\\
Tang \etal (2017)~\cite{Tang2017} & Yes & No\rlap{\tnote{4}} & Yes & No & 4.7 & Yes\\
Nguyen, Cao, and Jain (2018)~\cite{Nguyen2018a} & No\rlap{\tnote{2}} & No & Yes & No & 74.4 & Yes\\
Nguyen \etal (2020)~\cite{Nguyen2020} & Yes & No\rlap{\tnote{4}} & Yes & No & 7.0 & No\\
Zhou \etal (2020)~\cite{Zhou2020} & No\rlap{\tnote{2}} & No & Yes & No & 70.4 & No\\
Cao \etal (2020)~\cite{Cao2020} & No\rlap{\tnote{3}} & No & Yes & No & 5.5 & Yes\\
Zhang, Liu, and Liu (2021)~\cite{Zhang2021} & Yes & No\rlap{\tnote{4}} & Yes & No & 36.5 & No\\
Feng and Kumar (2023)~\cite{Feng2023} & Yes & No\rlap{\tnote{4}} & Yes & No & 2.2 & No\\
\addlinespace[0.25em]
LEADER (Ours) & \textbf{Yes} & \textbf{Yes} & \textbf{Yes} & \textbf{Yes} & \bfseries 0.9 & \textbf{Yes}\\
\bottomrule
\end{tabular*}
\begin{tablenotes}
\item[1] Sliding-window inference: requires multiple passes over local sub-regions.
\item[2] Two-stage pipeline: initial localization followed by region-specific refinement (e.g., ROI pooling or patch analysis).
\item[3] Multi-pass inference: five separate forward passes on externally enhanced inputs.
\item[4] Off-graph processing (e.g., post-inference NMS or angular decoding of minutiae directions).
\end{tablenotes}
\end{threeparttable}
\end{table*}

\section{The LEADER Architecture} 
\label{sec:method} 
The proposed architecture is built upon four key elements: (i)~cascaded feature extraction employing two encoder-decoder components with skip connections and attention-gating (Sec.~\ref{subsec:model});
(ii)~integrated postprocessing through on-graph NMS and trigonometric decoding (Sec.~\ref{subsec:postprocessing});
(iii)~adaptive ground-truth encoding for high-precision localization (Sec.~\ref{subsec:groundtruth});
and (iv)~multi-task optimization for synchronized attribute learning (Sec.~\ref{sec:loss}). For reference, the mathematical notation used throughout this work is provided in Sec.~\ref{subsec:notation}.

\subsection{Mathematical Notation}
\label{subsec:notation}
Table~\ref{tab:symbols} summarizes the mathematical notation used throughout this work, organized into three logical sections: (i)~\textit{Model and Inference}, covering input data, indexing, internal representations, and output data; (ii)~\textit{Ground-Truth-Encoding}, describing the topological supervision strategy; and (iii)~\textit{Training and Evaluation}, including the multi-task loss terms and the thresholds used for performance assessment. 

\begin{table}[t]
\centering
\setlength{\tabcolsep}{3pt}
\begin{threeparttable}
\caption{Mathematical symbols and notation.}
\label{tab:symbols}
\begin{tabular}{@{} c l l @{}}
\toprule
& \textbf{Symbol} & \textbf{Description} \\
\midrule

\multirow{16}{*}{\rotatebox[origin=c]{90}{\textit{Model \& Inference}}} 
& $\mathbf{F}$ & Raw input fingerprint image \\
& $w, h$ & Width and height of the input image \\
& $i, j$ & Row and column indices of an element in the spatial grid \\
& $\mathbf{X}, \mathbf{X}'$ & Generic intermediate feature maps \\
& $\mathbf{X}_{gate}$ & Recalibration gating signal within the \textit{Attention-Gate} \\
& $\varsigma(\cdot)$ & Sigmoid activation function \\
& $\psi(\cdot)$ & Point-wise convolution operator \\
& $[\cdot]$ & Channel-wise concatenation operator \\
& $\odot$ & Element-wise multiplication operator \\
& $\Delta_r$ & Dilated convolution with rate $r$ \\
& $\widehat{\mathbf{V}}_x, \widehat{\mathbf{V}}_y$ & Predicted Cartesian direction component maps \\
& $\widehat{\mathbf{P}}, \widehat{\mathbf{D}}, \widehat{\mathbf{T}}$ & Predicted maps for position, direction, and type \\
& $\widetilde{\mathbf{P}}$ & Predicted position map after on-graph NMS \\
& $\widehat{\mathbb{M}}, \hat{m}_k$ & Extracted minutiae set and $k$-th minutia \\
& $\hat{x}_k, \hat{y}_k$ & Coordinates of $\hat{m}_k$ \\
& $\hat{\theta}_k, \hat{t}_k, \hat{q}_k$ & Direction, type, and quality score of $\hat{m}_k$ \\
& $\tau_q$ & Confidence threshold for minutiae list extraction \\
\midrule

\multirow{10}{*}{\rotatebox[origin=c]{90}{\textit{Ground-Truth Encoding}}} 
& $\mathbb{M}, m_k$ & Ground-truth minutiae set and $k$-th minutia \\
& $x_k, y_k$ & Coordinates of $m_k$ \\
& $\theta_k, t_k$ & Direction and type of $m_k$ \\
& $\mathbf{P}, \mathbf{D}, \mathbf{T}$ & Ground-truth maps for position, direction, and type \\
& $\mathbf{W}$ & Spatial weight map used for loss balancing \\
& $\delta, \beta, \sigma, \lambda$ & CMR encoding hyperparameters \\
& $\omega(\cdot)$ & CMR piece-wise weighting function \\
& $\rho(\cdot)$ & Euclidean distance \\
& $\rho_{min}(\cdot)$ & Euclidean distance to the nearest positive pixel \\
& $\mathcal{G}(\cdot)$ & Gaussian function \\
\midrule

\multirow{8}{*}{\rotatebox[origin=c]{90}{\textit{Training \& Evaluation}}} 
& $\mathcal{L}$ & Composite multi-task loss function \\
& $\mathcal{L}_{p}, \mathcal{L}_{d}, \mathcal{L}_{t}$ & Position, directional and type loss terms \\
& $\alpha_p, \alpha_d, \alpha_t$ & Hyperparameters for multi-task loss balancing \\
& $\text{BCE}(\cdot)$ & Binary Cross-Entropy function \\
& $\phi(\cdot, \cdot)$ & Wrapping angular difference function \\
& $\epsilon$ & Small constant added for numerical stability \\
& $\mathcal{W}$ & Normalized weight magnitude (RMS of all parameters) \\
& $\rho_t, \theta_t$ & Thresholds for ground-truth pairing \\
\addlinespace[0.5em] 
\bottomrule
\end{tabular}
\end{threeparttable}
\end{table}

Tensors and matrices are denoted by bold uppercase letters (e.g., $\mathbf{P}, \mathbf{W}$), while their individual elements at row $i$ and column $j$ are represented by the corresponding lowercase italics (e.g., $p_{ij}, w_{ij}$). Sets are denoted by blackboard bold letters (e.g., $\mathbb{M}$), and tuples or generic scalars by lowercase italics (e.g., $m, r$).

\subsection{Cascaded Feature Extraction}
\label{subsec:model}
\begin{figure}[t]
    \centering
    \includegraphics[width=0.7\linewidth]{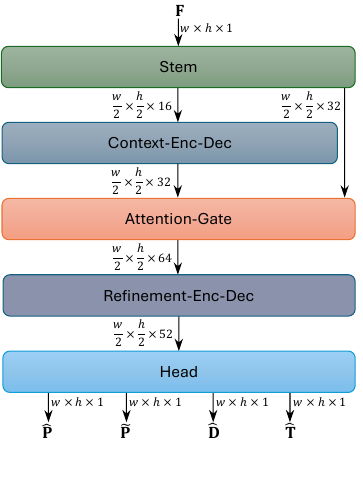}
    \caption{\textbf{High-level diagram of the LEADER model.} The network processes the fingerprint $\mathbf{F}$ through a pipeline composed of the \textit{Stem}, the \textit{Context-Enc-Dec}, the \textit{Attention-Gate}, the \textit{Refinement-Enc-Dec}, and the multi-task \textit{Head} producing position ($\widehat{\mathbf{P}}, \widetilde{\mathbf{P}}$), direction ($\widehat{\mathbf{D}}$), and type ($\widehat{\mathbf{T}}$) maps.}
    \label{fig:model}
\end{figure}

LEADER comprises five cascaded components (Fig.~\ref{fig:model}) designed to balance spatial precision with semantic depth. Detailed diagrams of each component and their constituent functional blocks are provided in the Supplementary Material (Figs.~\ref{fig:supp_stem}--\ref{fig:supp_blocks}).

\subsubsection{Stem}
This component performs initial feature extraction and spatial mapping, accepting the fingerprint $\mathbf{F} \in \mathbb{R}^{w \times h \times 1}$ as input.\footnote{To ensure compatibility with downsampling layers, $\mathbf{F}$ is padded to dimensions that are multiples of 32 pixels, using the intensity of the top-left pixel as a constant background value.} The \textit{Stem} adopts a dual-path configuration (Fig.~\ref{fig:supp_stem}) comprising two \textit{StemBlocks} with distinct pooling strategies. 
Each \textit{StemBlock} (Fig.~\ref{fig:supp_blocks}d) is designed to capture local ridge details and wider contextual patterns.

\subsubsection{Context-Enc-Dec}
This stage extracts mid-level topological features through a symmetric encoder-decoder with skip connections (Fig.~\ref{fig:supp_sae}). The encoding path employs \textit{SeparableConvBlocks} (Fig.~\ref{fig:supp_blocks}c) for computationally efficient feature extraction, while spatial resolution is progressively reduced through \textit{DownsamplingBlocks} (Fig.~\ref{fig:supp_blocks}a) to attenuate noise and expand the effective receptive field. Symmetrically, the decoding path restores the original resolution using \textit{UpsamplingBlocks} (Fig.~\ref{fig:supp_blocks}b). Feature maps from the encoder are transferred to the decoder through channel-wise concatenation (skip connections), ensuring the preservation of fine-grained structural details alongside high-level representations.

\subsubsection{Attention-Gate}
To prioritize salient ridge structures, this component performs multi-scale and channel-wise recalibration of the feature maps $\mathbf{X}$ generated by the \textit{Context-Enc-Dec}. The gating mechanism (Fig.~\ref{fig:supp_ga}) captures contextual information across varying receptive-field sizes through three parallel convolutional paths with different dilation rates. Formally, the gating signal $\mathbf{X}_{gate}$ is defined as $\mathbf{X}_{gate} = \varsigma( \psi( [\Delta_1(\mathbf{X}), \Delta_3(\mathbf{X}), \Delta_6(\mathbf{X})] ) )$, where $[\cdot]$ denotes channel-wise concatenation, $\Delta_r$ represents a $3\times3$ convolution with 16 filters and dilation rate $r \in \{1, 3, 6\}$ followed by a GELU activation, $\psi$ is a point-wise convolution that reduces the channel depth to 32, and $\varsigma$ is the Sigmoid activation function. The refined feature maps $\mathbf{X}'$ are obtained by modulating the original features $\mathbf{X}$ through element-wise multiplication with the gating signal, $\mathbf{X}' = \mathbf{X} \odot \mathbf{X}_{gate}$. These are subsequently concatenated with high-resolution features from the \textit{Stem} via a direct skip connection (Fig.~\ref{fig:model}). This design ensures that the final refinement stage operates on a saliency-calibrated representation spatially aligned to the original ridge geometry.

\subsubsection{Refinement-Enc-Dec}
This stage (Fig.~\ref{fig:supp_sae}) performs high-level semantic refinement by replacing separable convolutions of the \textit{Context-Enc-Dec} with \textit{InvBottleneckConvBlocks} (Fig.~\ref{fig:supp_blocks}f). These blocks employs an expansion-contraction pattern, projecting features into a higher-dimensional space through point-wise convolutions before compressing them to the original depth. This design facilitates complex representation learning while maintaining high parameter efficiency.
Furthermore, a non-monotonic channel distribution enforces a strategic bottleneck: after reaching a peak depth of 128 channels, the network sharply contracts to 32 channels at the fourth block and 27 at the subsequent one. This localized compression at the transition between the contracting and expanding paths is designed to filter redundant information while preserving essential topology. The remainder of the decoder is optimized with precisely tuned channel dimensions to maintain representational capacity and an overall footprint of 0.9\,M parameters.

\subsubsection{Head}
This component transforms the feature maps generated by the \textit{Refinement-Enc-Dec} into the final minutiae attributes. Its structure (Fig.~\ref{fig:supp_head}) branches into three parallel \textit{HeadBlocks} (Fig.~\ref{fig:supp_blocks}e), each employing an \textit{InvBottleneckConvBlock} with six filters. The \textit{Head} decodes the 52 input feature maps into task-specific projections through specialized output layers, using Sigmoid activations for the position and type maps ($\widehat{\mathbf{P}}, \widehat{\mathbf{T}}$), and linear regression for the direction map $\widehat{\mathbf{D}}$ to maintain its dynamic range. Additionally, an integrated on-graph postprocessing step applies non-maximum suppression to $\widehat{\mathbf{P}}$ to yield the refined position map $\widetilde{\mathbf{P}}$, as detailed below.

\subsection{Integrated Postprocessing}
\label{subsec:postprocessing}
The \textit{Head} component internalizes the extraction logic typically performed by off-graph heuristics, producing four synchronized output maps directly within the computational flow.

\subsubsection{On-graph NMS}
A specific deterministic, parameter-free layer (Fig.~\ref{fig:supp_head}, Supplementary Material) replaces external peak-finding algorithms. The predicted position map $\widehat{\mathbf{P}}$ is smoothed with a $5 \times 5$ Gaussian kernel to yield $\widehat{\mathbf{P}}^\ast$. The refined map $\widetilde{\mathbf{P}}$ is then generated through local spatial competition:
\begin{equation}
    \widetilde{p}_{ij} = \begin{cases} 
    \hat{p}^\ast_{ij} & \text{if } \hat{p}^\ast_{ij} = \max_{(x,y) \in \mathcal{N}_7(j,i)} \hat{p}^\ast_{yx} \\
    0 & \text{otherwise}
    \end{cases}
\end{equation}
where $\mathcal{N}_7(j,i)$ denotes a $7 \times 7$ spatial neighborhood centered at $(j,i)$. This operation is implemented through unit-stride MaxPooling2D followed by element-wise comparison.

\subsubsection{Trigonometric Directional Decoding}
To account for the circular nature of minutiae directions while maintaining optimization stability, the network regresses two Cartesian component maps, $(\widehat{\mathbf{V}}_x, \widehat{\mathbf{V}}_y)$, rather than a single angular representation. This approach maps the targets into a continuous Euclidean space, avoiding the discontinuities and instability inherent in periodic angular regression. The final direction map $\widehat{\mathbf{D}}$ is subsequently recovered on-graph by the deterministic, parameter-free \textit{CartesianToPolar} layer (Fig.~\ref{fig:supp_head}, Supplementary Material) using the element-wise differentiable \textit{arctan2} operator: $\widehat{\mathbf{D}}=\operatorname{arctan2}(\widehat{\mathbf{V}}_{y},\widehat{\mathbf{V}}_{x})$.

\subsubsection{Minutiae List Extraction}
\label{subsec:minutiae_list_readout}
The dense output maps are converted into a sparse minutiae set $\widehat{\mathbb{M}}$ through a straightforward, non-heuristic readout pass. For every pixel where $\tilde{p}_{ij} \ge \tau_q$, a minutia tuple $\hat{m}_k = (\hat{x}_k, \hat{y}_k, \hat{\theta}_k, \hat{t}_k, \hat{q}_k)$ is added to $\widehat{\mathbb{M}}$, where:
\begin{itemize}
    \item $(\hat{x}_k, \hat{y}_k) = (j, i)$ are the Cartesian coordinates; 
    \item $\hat{\theta}_k = \hat{d}_{ij}$ is the predicted direction; 
    \item $\hat{t}_k$ is the type: ridge ending if $\hat{t}_{ij} \ge 0.5$, bifurcation otherwise;
    \item $\hat{q}_k = \tilde{p}_{ij}$ is a reliability value representing the quality score of $\hat{m}_k$.
\end{itemize}

\subsection{Adaptive Ground-Truth Encoding}
\label{subsec:groundtruth}
A critical challenge in training deep networks for minutiae extraction lies in the translation of discrete ground-truth annotations into continuous heatmaps suitable for multi-task learning. Direct regression of sparse attributes for each minutia $m_k = (x_k, y_k, \theta_k, t_k)$ is ill-posed due to extreme class imbalance and the spatial uncertainty of manual annotations~\cite{Malhotra2021}.

Recent deep learning methods have increasingly prioritized this encoding step. Early approaches relied on coarse-grained heatmaps at $w/8 \times h/8$ resolution~\cite{Tang2017}, whereas subsequent approaches adopted $w/4 \times h/4$ representations~\cite{Zhou2020} and incorporated Gaussian kernels to produce smoother heatmaps~\cite{Zhang2021}. More recently, a concentric-zone encoding strategy was introduced~\cite{Feng2023}, defining positive and negative regions around each minutia separated by a neutral annular buffer; pixels within this transition zone are specifically excluded from the loss function to mitigate gradient ambiguity.

\begin{figure}[t]
    \centering
    \includegraphics[width=\linewidth]{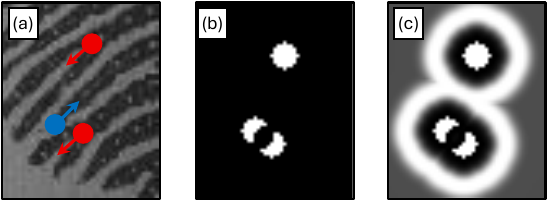}
    \caption{\textbf{Ground-truth encoding and adaptive spatial weighting.} (a) Detail of a fingerprint image with annotated minutiae. (b) Generated binary position map $\mathbf{P}$: the morphology of the positive components is dynamically constrained in high-density regions to prevent spatial overlap. (c) Corresponding weight map $\mathbf{W}$, derived from the CMR geometry. The penalty zones adaptively reshape to maintain separation between closely spaced minutiae; the \textit{castle} regions in $\mathbf{W}$ correspond to positive pixels in $\mathbf{P}$, while the \textit{ramparts} provide localized penalty peaks to enforce sharp localization.}
    \label{fig:target_maps}
\end{figure}

Building upon these concepts, LEADER introduces a more sophisticated encoding and weighting strategy designed to preserve localization precision in high-density regions. Unlike static approaches, this method employs an adaptive formulation that dynamically reshapes target regions based on minutiae proximity. The binary ground-truth minutiae position map $\mathbf{P}$ is defined as:
\begin{equation}
    p_{ij} = \begin{cases} 
    1 & \text{if } \exists! m_k \in \mathbb{M} : \rho(m_k, i, j) \le \delta \\
      & \phantom{if }\text{and } \nexists m_k \in \mathbb{M} : \delta < \rho(m_k, i, j) \le \delta + \beta \\
    0 & \text{otherwise}
    \end{cases}
\end{equation}
where $\rho(m_k, i, j) = \sqrt{(j - x_k)^2 + (i - y_k)^2}$ denotes the Euclidean distance between minutia $m_k$ and the pixel at Cartesian coordinates $(j, i)$.
Under this definition, each connected component of pixels labeled as 1 is unambiguously associated with exactly one minutia. In regions of high minutiae density, this logic dynamically constrains the positive regions to ensure a separation of at least $\beta$ pixels between neighboring features. This results in non-circular morphologies (e.g., those observed in Fig. \ref{fig:target_maps}b), which effectively prevent gradient interference during training.

To guide the optimization, a spatial weight map $\mathbf{W}$ is derived, where each element is computed as $w_{ij}=\omega\left(\rho_{min}(\mathbf{P}, i, j)\right)$. Here, $\rho_{min}(\mathbf{P}, i, j)$ represents the Euclidean distance to the nearest positive pixel in $\mathbf{P}$:
\begin{equation}
    \rho_{min}(\mathbf{P}, i, j) = \min_{(x,y) :  p_{yx}=1} \sqrt{(j-x)^2 + (i-y)^2}
\end{equation}

\begin{figure}[t]
    \centering
    \includegraphics[width=\linewidth]{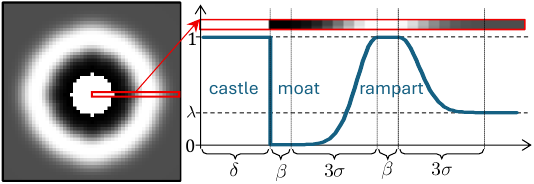}
    \caption{\textbf{Radial profile of the CMR weighting strategy.} Left: 2D weight map $\mathbf{W}$ for an isolated minutia. Right: Profile of the weighting function $\omega(\rho_{\min}(\cdot))$ along a cross-section from the center (pixels are highlighted in red). The \textit{castle} ($w_{ij}=1$) covers the central region—nominally of radius $\delta$, but dynamically adjusted in high-density areas—followed by a zero-gradient \textit{moat} of width $\beta$, which effectively extends into the lower tail of the Gaussian slope. The \textit{rampart} rises to a peak penalty ($w_{ij}=1$) to enforce sharp localization before descending to the background plateau $\lambda$.}
    \label{fig:weight_plot}
\end{figure}

The weight map is modulated by a piece-wise function $\omega(s)$, designed to generate the CMR geometry (Fig.~\ref{fig:weight_plot}):
\begin{equation}
    \omega(s) = \begin{cases} 
    1 & \text{if } s = 0 \\
    0 & \text{if } 0 < s \le \beta \\
    \mathcal{G}(s - s_1) & \text{if } \beta < s \le s_1 \\
    1 & \text{if } s_1 < s \le s_1 + \beta \\
    \lambda + (1 - \lambda)\mathcal{G}(s - s_1 - \beta) & \text{otherwise}
    \end{cases}
\end{equation}
where $s_1 = \beta + 3\sigma$ and $\mathcal{G}(z) = \exp(-z^2/2\sigma^2)$. As illustrated in Fig. \ref{fig:weight_plot}, this weighting function yields a profile consisting of a positive core (\textit{castle}), a zero-gradient uncertainty buffer (\textit{moat}) to mitigate labeling jitter, and a localized penalty peak (\textit{rampart}) that enforces sharp localization. This geometry ensures that the network prioritizes accurate placement by penalizing false activations in the immediate vicinity of each minutia before the weight descends to the background plateau $\lambda$. In high-density areas, the interaction of these adaptive profiles ensures that the network resolves closely spaced minutiae without interference (Fig.~\ref{fig:target_maps}c).

Finally, the ground-truth direction map $\mathbf{D}$ and type map $\mathbf{T}$ are generated by assigning the attributes of the nearest minutia $m^* \in \mathbb{M}$ to each pixel $(i,j)$:
\begin{equation}
d_{ij} = \theta^*, \quad t_{ij} = t^*, \quad \text{where } m^* = \operatorname*{arg\,min}_{m_k \in \mathbb{M}} \rho(m_k, i, j)
\end{equation}
where $\theta^*$ and $t^*$ denote the direction and type of $m^*$, respectively. While these maps are defined globally for implementation convenience, they are subject to a conditional masking strategy during training; specifically, the loss terms for direction and type are restricted to pixels where $p_{ij}=1$ to ensure that gradients are derived exclusively from valid minutiae locations (see Sec.~\ref{sec:loss}).

The CMR encoding is governed by four hyperparameters ($\delta, \beta, \sigma, \lambda$), tuned for 500\,DPI fingerprint sensors. Further details are provided in Appendix~\ref{sec:training} (Supplementary Material).

\subsection{Multi-task Optimization}
\label{sec:loss}
LEADER is trained by means of a composite loss function $\mathcal{L} = \alpha_p \mathcal{L}_{p} + \alpha_d \mathcal{L}_{d} + \alpha_t \mathcal{L}_{t}$, where hyperparameters $\alpha_{p}$, $\alpha_{d}$, and $\alpha_{t}$ balance spatial localization, directional regression, and type classification, respectively. Each loss term ($\mathcal{L}_{p}$, $\mathcal{L}_{d}$, and $\mathcal{L}_{t}$) is computed by comparing the predicted maps ($\widehat{\mathbf{P}}, \widehat{\mathbf{D}}, \widehat{\mathbf{T}}$) with their ground-truth counterparts ($\mathbf{P},\mathbf{D},\mathbf{T}$). Notably, the NMS-refined map $\widetilde{\mathbf{P}}$ is excluded from optimization; as a deterministic transformation of $\widehat{\mathbf{P}}$ through non-trainable layers, its role is to provide an integrated inference path rather than a supervisory target.

The position loss $\mathcal{L}_{p}$ is defined as a weighted Binary Cross Entropy (BCE) modulated by the CMR weight map $\mathbf{W}$:
\begin{equation}
    \mathcal{L}_{p} = \frac{\sum_{i,j} w_{ij} \cdot \text{BCE}(p_{ij}, \hat{p}_{ij})}{\sum_{i,j} w_{ij} + \epsilon}
\end{equation}
This formulation ensures a strong supervisory signal at true minutiae locations (\textit{castle}) and enforces sharp localization via the \textit{rampart} penalty, while the zero-weight \textit{moat} prevents ambiguous gradients from hindering the convergence in regions of high labeling uncertainty.

Secondary attributes are supervised through a conditional masking strategy. Since predicting direction or type at non-minutia locations is ill-posed, these pixels are excluded from the loss via $\mathbf{P}$.
The directional loss $\mathcal{L}_{d}$ is computed as the normalized masked Root Mean Square (RMS) of the angular difference $\phi(d_{ij}, \hat{d}_{ij}) = [ (d_{ij} - \hat{d}_{ij} + \pi) \bmod 2\pi ] - \pi$:
\begin{equation}
    \mathcal{L}_{d} = \frac{1}{\pi} \sqrt{ \frac{\sum_{i,j} p_{ij} \cdot \phi(d_{ij}, \hat{d}_{ij})^2}{\sum_{i,j} p_{ij} + \epsilon} }
\end{equation}
Similarly, the type loss $\mathcal{L}_{t}$ is calculated as the masked BCE between $\mathbf{T}$ and $\widehat{\mathbf{T}}$:
\begin{equation}
    \mathcal{L}_{t} = \frac{\sum_{i,j} p_{ij} \cdot \text{BCE}(t_{ij}, \hat{t}_{ij})}{\sum_{i,j} p_{ij} + \epsilon}
\end{equation}

\section{Experimental Results}
\label{sec:experiments}
This section evaluates LEADER, demonstrating that a lightweight network can achieve top-tier performance across diverse fingerprint modalities using a single, fixed set of weights. We assess the model’s accuracy, efficiency, and generalization through the following experimental structure: Sec.~\ref{subsec:setup} describes the datasets and evaluation protocol, followed by a comparative analysis against established methods on both plain and latent fingerprints (Sec.~\ref{subsec:sota_comparison}). An ablation study then isolates the contributions of key structural components alongside the proposed CMR weighting strategy (Sec.~\ref{subsec:ablation}). Finally, Sec.~\ref{subsec:interpretability} provides a qualitative visual analysis to investigate the internal representations learned by the model.
To ensure full reproducibility, comprehensive details regarding data augmentation, optimizer schedules, and hyperparameters---including an analysis of the training dynamics---are provided in Appendix~\ref{sec:training} of the Supplementary Material.

\subsection{Experimental Setup}
\label{subsec:setup}
LEADER is trained and evaluated using a heterogeneous collection of fingerprint datasets (Table~\ref{tab:datasets}, Fig.~\ref{fig:datasets}), encompassing a wide spectrum of acquisition technologies and noise patterns characteristic of real-world applications.

\begin{table*}[htb]
\centering
\begin{threeparttable}
\caption{Characteristics of the datasets used for training, hyperparameter tuning, and testing.}
\label{tab:datasets}
\begin{tabular*}{\textwidth}{@{\extracolsep{\fill}} l l S[table-format=4.0] l S[table-format=5.0] r @{}}
\toprule
\textbf{Dataset} & \textbf{Sensor Type} & \textbf{Images} & \textbf{Ground-truth Source}\tnote{1} & \textbf{Minutiae} & \textbf{Usage} \\
\midrule
FVC2002 DB2-A & Optical & 100\rlap{\tnote{2}} & Manual (Authors) & 5041 & \multirow{6}{*}{Training} \\
FVC2002 DB3-A & Capacitive & 800 & \cite{Thai2016} \cite{Kayaoglu2013} & 19032 \\
FVC2004 DB1-A & Optical & 800 & \cite{Thai2016} \cite{Kayaoglu2013} & 32771 \\
FVC2004 DB3-A & Thermal Sweeping & 800 & \cite{Thai2016} \cite{Kayaoglu2013} & 32608 \\
FFE Good & Optical & 10 & \cite{Cappelli2024a}\tnote{3} & 525 \\
FFE Bad & Optical & 50 & \cite{Cappelli2024a}\tnote{3} & 2107 \\
\midrule
FVC2000 DB2-A & Capacitive & 100\rlap{\tnote{4}} & Manual (Authors) & 4298 & Tuning \\
\midrule
FVC2002 DB1-A & Optical & 100\rlap{\tnote{4}} & \cite{Thai2016} \cite{Kayaoglu2013} & 4071 & \multirow{3}{*}{Testing} \\
NIST SD27 & Latent & 258 & \cite{Feng2013} \cite{Garris2000} & 5303 \\
NIST SD302 & Latent & 3815 & \cite{Fiumara2026} & 82793 \\
\bottomrule
\end{tabular*}
\begin{tablenotes}
\item[1] When two references are cited, the first refers to the segmentation mask and the second to the minutiae ground-truth.
\item[2] Only the third impression of each of the 100 fingers was used.
\item[3] FFE ground-truth minutiae were extracted from manually-labeled ridge skeletons using the traditional crossing number method.
\item[4] Only the first impression of each of the 100 fingers was used. 
\end{tablenotes}
\end{threeparttable}
\end{table*}

\subsubsection{Training and Tuning Datasets}
The training set comprises 2,560 plain fingerprints derived from six distinct sources, containing a total of 92,084 ground-truth minutiae. We include subsets from the FVC2002~\cite{Maio2002a} and FVC2004~\cite{Cappelli2006} benchmarks to expose the network to diverse ridge patterns and sensor-specific artifacts. Specifically, FVC2002 DB2-A and DB3-A introduce variations in skin moisture and exaggerated finger rotation, while the FVC2004 subsets (DB1-A, DB3-A) present severe skin distortions, varying contact pressure, and noise from excessively wet or dry fingers. FVC2004 DB3-A (thermal sweeping sensor) is particularly challenging due to reconstruction artifacts. To enhance the model's capacity to handle fine structural details in both high- and low-quality fingerprints, the FFE Good and FFE Bad datasets from~\cite{Cappelli2024a} are incorporated: the former contains high-quality samples, while the latter focuses on fingerprints affected by cuts, abrasions, and severe skin-moisture variations. Ground-truth segmentation masks for training samples are employed exclusively during data augmentation, as detailed in Appendix~\ref{sec:training} (Supplementary Material).

Hyperparameters, including those governing the CMR encoding strategy (see Sec.~\ref{subsec:groundtruth}), are tuned using the first impression of each finger in FVC2000 DB2-A~\cite{Maio2002}. This dataset consists of fingerprints acquired from untrained users without quality control, providing a representative baseline for parameter optimization. All experimental evaluations across both plain and latent benchmarks are conducted using a single, unified model instance trained with these fixed hyperparameters, without any dataset-specific fine-tuning or re-calibration, while the confidence threshold $\tau_q$ is dynamically swept to determine the peak operating point per dataset—following a standardized evaluation protocol applied uniformly across all methods. The model operates at the standard resolution of 500\,DPI.

\subsubsection{Test Datasets}
The performance is evaluated on three held-out test sets\footnote{Regarding dataset separation, there is strictly zero image overlap between the training, tuning, and testing sets. Furthermore, the two primary evaluation benchmarks (NIST SD27 and NIST SD302) originate from US collections, ensuring complete independence in subjects, acquisition environments, and sensor modalities from the European-collected training datasets. Finally, training minutiae ground truth relies on author curation and literature benchmarks~\cite{Kayaoglu2013}, whereas the ground truths for both latent benchmarks were independently annotated by certified forensic examiners~\cite{Garris2000, Fiumara2026}, preventing annotation-bias leakage.} to assess generalization capabilities. The first, FVC2002 DB1-A~\cite{Maio2002a}, represents a standard evaluation benchmark for plain optical fingerprints. The second, NIST SD27~\cite{Garris2000}, features latent fingerprints recovered from crime scenes. Minutiae for this set were validated by professional latent examiners~\cite{Garris2000}, while segmentation masks were sourced from~\cite{Feng2013}. The third test dataset is NIST SD302~\cite{Fiumara2019}, specifically utilizing the annotated latent distal phalanx subset (SD302h)~\cite{Fiumara2026}. These latent impressions were left naturally by subjects during simulated daily activities (e.g., handling objects or paper currency) and subsequently processed using standard forensic development techniques~\cite{Fiumara2019}. To maintain consistency with the other training and evaluation benchmarks, the image resolution was downsampled from its native 1000\,DPI to 500\,DPI. The dataset consists of the 3,815 latent images categorized as ``value'' for forensic comparison by expert analysis. Both the segmentation ground truth masks and the minutiae were annotated and validated by certified latent print examiners~\cite{Fiumara2026}. Together, the two NIST latent datasets present the most significant recognition challenges, characterized by heavy background clutter, partial ridge structures, overlapping patterns, and severe environmental noise.

\begin{nofigurediff*}[htb]
    \centering
    \includegraphics[width=\linewidth]{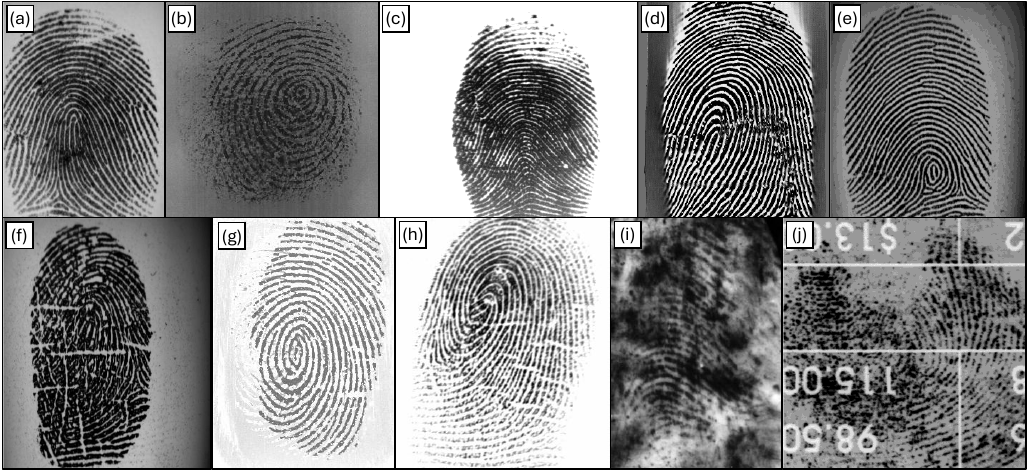}
    \caption{\textbf{Representative fingerprint samples from the datasets.} 
    The collection covers diverse acquisition technologies: 
    (a)~FVC2002 DB2-A (Optical); 
    (b)~FVC2002 DB3-A (Capacitive); 
    (c)~FVC2004 DB1-A (Optical); 
    (d)~FVC2004 DB3-A (Thermal sweeping); 
    (e)~FFE Good (Optical); 
    (f)~FFE Bad (Optical); 
    (g)~FVC2000 DB2-A (Capacitive); 
    (h)~FVC2002 DB1-A (Optical); 
    (i)~NIST SD27 (Latent);
    (j)~NIST SD302 (Latent).}
    \label{fig:datasets}
\end{nofigurediff*}

\subsubsection{Evaluation Metrics and Protocol}
Performance in minutiae extraction is typically assessed through either \textit{intrinsic evaluation}, which compares extracted minutiae directly against ground truth, or \textit{extrinsic evaluation} based on recognition accuracy (e.g., EER). This work adopts the former to decouple performance from the biases of minutiae-comparison algorithms, which may mask extractor inaccuracies (such as errors in position, direction, or type) via error-tolerant heuristics.
The evaluation follows a systematic protocol based on four steps:
\begin{enumerate}[label=(\roman*)]
    \item \textit{Minutiae pairing and filtering}---Extracted and ground-truth minutiae are paired by solving the assignment problem~\cite{Crouse2016}. A match is defined as a True Positive (TP) if the Euclidean distance $\rho(\cdot)$ and angular difference $\phi(\cdot)$ are within thresholds $\rho_t$ and $\theta_t$, respectively. We adopt three increasingly stringent levels: $(\rho_t=16\text{\,px}, \theta_t=\pi/6\text{\,rad})$, $(12, \pi/8)$, and $(8, \pi/10)$. To ensure an objective and isolated evaluation of minutiae extraction capability, all fingerprints are cropped to the ridge-area bounding box defined by ground-truth masks, avoiding confounding biases from upstream automatic segmentation errors. Furthermore, minutiae located outside the ground-truth mask or within a 14\,px margin of the segmentation boundary are filtered out. While LEADER is inherently robust to boundary-induced spurious detections, this filtering is applied uniformly to avoid penalizing competitors for easily mitigable artifacts.\footnote{Sensitivity analysis removing this boundary filtering  confirms that relative method rankings remain stable, see Appendix~\ref{sec:border_sens_analysis} (Supplementary Material).}
    \item \textit{Aggregate performance computation}---Based on the pairing results, we report Precision-Recall (PR) curves and the $F_1$-score at the optimal operating point. For methods providing continuous quality scores (including LEADER), full PR curves are generated, and the peak $F_1$-score is identified by sweeping the confidence threshold across each dataset.\footnote{Operating the evaluation at dataset-specific optimal thresholds ensures a fair, unbiased assessment of each extractor's maximum intrinsic capability, preventing competitors from being artificially penalized by default factory thresholds typically calibrated for plain fingerprints.} For methods providing unscored minutiae outputs, their single operating point is evaluated directly. Both \textit{type-agnostic} and \textit{type-aware} regimes are considered, where the latter requires matching minutia types for a TP.
    \item \textit{Sample-level ranking}---To assess consistency, extractors are ranked by the $F_1$-score achieved on each fingerprint, computed at the previously identified optimal operating point. We report the mean rank (with standard deviation) and the frequency of achieving top or top-3 positions. Furthermore, the \textit{bottom half percentage} tracks how often a method falls into the lower 50\% of the evaluation pool, serving as a proxy for reliability. Finally, a pairwise \textit{direct-win} analysis quantifies how often a method outperforms each specific competitor on a per-sample basis.
    \item \textit{Efficiency analysis}---Computational cost is evaluated in terms of execution latency (on both CPU and GPU) to assess suitability for real-time applications.
\end{enumerate}

\newcommand{\ci}[2]{\raisebox{0.3ex}{\scalebox{0.85}{\tiny$\substack{\hfill #1\\[-0.1ex]\hfill #2}$}}}
\newcommand{\cisym}[1]{\raisebox{0.3ex}{\scalebox{0.85}{\tiny$\pm #1$}}}
\newcommand{\thrl}[2]{\ensuremath{\substack{#1\,\text{px}\\\pi/#2\,\text{rad}}}}

\begin{notablediff*}[ht]
\centering
\begin{threeparttable}
\caption{Comparative performance ($F_1$-score) across the three threshold levels under type-agnostic and type-aware regimes.}
\label{tab:results_f1}
\begin{tabular}{l l ccc c ccc c ccc} 
\toprule
& & \multicolumn{3}{c}{\textbf{FVC2002 DB1-A}} & & \multicolumn{3}{c}{\textbf{NIST SD27}} & & \multicolumn{3}{c}{\textbf{NIST SD302}} \\
\cmidrule(lr){3-5} \cmidrule(lr){7-9}  \cmidrule(lr){11-13}
& \textbf{Method} & \thrl{16}{6} & \thrl{12}{8} & \thrl{8}{10} &
                  & \thrl{16}{6} & \thrl{12}{8} & \thrl{8}{10} &
                  & \thrl{16}{6} & \thrl{12}{8} & \thrl{8}{10} \\
\midrule
\multirow{10}{*}{\rotatebox[origin=c]{90}{\textit{Type-agnostic}}} 

& FingerNet & \underline{0.87}\cisym{0.01} & \underline{0.86}\ci{+0.01}{-0.02} & \underline{0.83}\cisym{0.02} & & \underline{0.67}\cisym{0.02} & \underline{0.64}\cisym{0.02} & \underline{0.57}\cisym{0.02} & & \underline{0.65}\cisym{0.01} & \underline{0.64}\cisym{0.01} & \underline{0.60}\cisym{0.01} \\
& LatentAFIS & 0.84\cisym{0.01} & 0.83\cisym{0.01} & 0.81\cisym{0.01} & & 0.62\cisym{0.02} & 0.59\cisym{0.02} & 0.54\cisym{0.02} & & 0.57\cisym{0.01} & 0.56\cisym{0.01} & 0.52\cisym{0.01} \\
& MinutiaeNet & 0.47\cisym{0.02} & 0.43\cisym{0.02} & 0.38\cisym{0.02} & & 0.48\cisym{0.02} & 0.45\cisym{0.02} & 0.40\cisym{0.02} & & 0.25\cisym{0.01} & 0.21\cisym{0.01} & 0.17\cisym{0.01} \\
\addlinespace[0.25em]
& FingerJet & 0.83\ci{+0.01}{-0.02} & 0.80\cisym{0.02} & 0.75\cisym{0.02} & & 0.40\cisym{0.03} & 0.36\cisym{0.03} & 0.29\cisym{0.03} & & 0.22\cisym{0.01} & 0.21\cisym{0.01} & 0.18\cisym{0.01} \\
& FDx & 0.86\cisym{0.01} & 0.85\cisym{0.01} & 0.81\cisym{0.02} & & 0.45\cisym{0.03} & 0.41\cisym{0.03} & 0.34\cisym{0.03} & & 0.39\cisym{0.01} & 0.36\cisym{0.01} & 0.33\cisym{0.01} \\
& VeriFinger & 0.86\cisym{0.01} & 0.85\cisym{0.01} & 0.80\cisym{0.01} & & 0.64\cisym{0.02} & 0.61\cisym{0.02} & 0.53\ci{+0.02}{-0.03} & & 0.60\cisym{0.01} & 0.57\cisym{0.01} & 0.52\cisym{0.01} \\
\addlinespace[0.25em]
& MINDTCT & 0.84\cisym{0.01} & 0.82\ci{+0.01}{-0.02} & 0.78\cisym{0.02} & & 0.32\cisym{0.03} & 0.29\cisym{0.03} & 0.24\cisym{0.03} & & 0.30\cisym{0.01} & 0.28\cisym{0.01} & 0.25\cisym{0.01} \\
& FingerFlow & 0.53\cisym{0.02} & 0.49\cisym{0.02} & 0.43\cisym{0.02} & & 0.49\cisym{0.02} & 0.47\cisym{0.02} & 0.42\cisym{0.02} & & 0.26\cisym{0.01} & 0.22\cisym{0.01} & 0.18\cisym{0.01} \\
& SourceAFIS & 0.78\cisym{0.01} & 0.76\cisym{0.02} & 0.72\cisym{0.02} & & 0.29\cisym{0.02} & 0.25\cisym{0.02} & 0.19\cisym{0.02} & & 0.28\cisym{0.01} & 0.24\cisym{0.01} & 0.20\cisym{0.01} \\
\addlinespace[0.25em]
& LEADER (Ours) & \textbf{0.92}\cisym{0.01} & \textbf{0.91}\cisym{0.01} & \textbf{0.90}\cisym{0.01} & & \textbf{0.71}\cisym{0.02} & \textbf{0.69}\cisym{0.02} & \textbf{0.62}\cisym{0.02} & & \textbf{0.69}\cisym{0.01} & \textbf{0.67}\cisym{0.01} & \textbf{0.64}\cisym{0.01} \\
\midrule
\multirow{7}{*}{\rotatebox[origin=c]{90}{\textit{Type-aware}}}
& FingerJet & 0.64\cisym{0.02} & 0.62\cisym{0.02} & 0.58\cisym{0.02} & & 0.27\cisym{0.02} & 0.25\cisym{0.02} & 0.20\cisym{0.02} & & 0.16\cisym{0.01} & 0.15\cisym{0.01} & 0.13\cisym{0.01} \\
& FDx & \underline{0.73}\cisym{0.02} & \underline{0.71}\cisym{0.02} & \underline{0.69}\cisym{0.02} & & 0.30\cisym{0.02} & 0.28\cisym{0.02} & 0.24\cisym{0.02} & & 0.27\cisym{0.01} & 0.25\cisym{0.01} & 0.23\cisym{0.01} \\
& VeriFinger & 0.71\cisym{0.02} & 0.70\cisym{0.02} & 0.67\cisym{0.02} & & \underline{0.42}\cisym{0.02} & \underline{0.39}\cisym{0.02} & \underline{0.35}\cisym{0.02} & & \underline{0.41}\cisym{0.01} & \underline{0.39}\cisym{0.01} & \underline{0.36}\cisym{0.01} \\
\addlinespace[0.25em]
& MINDTCT & 0.72\cisym{0.02} & 0.70\cisym{0.02} & 0.66\cisym{0.02} & & 0.21\cisym{0.03} & 0.19\cisym{0.03} & 0.17\cisym{0.02} & & 0.22\cisym{0.01} & 0.20\cisym{0.01} & 0.18\cisym{0.01} \\
& FingerFlow & 0.34\cisym{0.02} & 0.31\cisym{0.02} & 0.27\cisym{0.02} & & 0.29\cisym{0.02} & 0.27\cisym{0.02} & 0.25\cisym{0.02} & & 0.17\cisym{0.01} & 0.14\cisym{0.01} & 0.12\cisym{0.01} \\
& SourceAFIS & 0.62\cisym{0.02} & 0.60\cisym{0.02} & 0.58\cisym{0.02} & & 0.19\cisym{0.02} & 0.17\cisym{0.02} & 0.13\cisym{0.01} & & 0.18\cisym{0.01} & 0.16\cisym{0.01} & 0.13\cisym{0.01} \\
\addlinespace[0.25em]
& LEADER (Ours) & \textbf{0.85}\ci{+0.01}{-0.02} & \textbf{0.84}\cisym{0.01} & \textbf{0.83}\cisym{0.01} & & \textbf{0.52}\ci{+0.03}{-0.02} & \textbf{0.51}\ci{+0.02}{-0.03} & \textbf{0.47}\cisym{0.03} & & \textbf{0.53}\cisym{0.01} & \textbf{0.52}\cisym{0.01} & \textbf{0.50}\cisym{0.01} \\

\bottomrule
\end{tabular}
\begin{tablenotes}
\item  Nominal scores reflect full test sets, with 95\% confidence intervals estimated via 1000 paired non-parametric image-level bootstrap resamples. Symmetric bounds at two-decimal precision are shown as $\pm$, while asymmetric bounds are formatted as $_{-low}^{+high}$.
\item \textbf{Bold} and \underline{underlined} values indicate best and second-best scores, respectively.
\end{tablenotes}
\end{threeparttable}
\end{notablediff*}

\subsection{Performance Evaluation and Comparative Analysis}
\label{subsec:sota_comparison}
To ensure a rigorous and transparent evaluation, LEADER is compared against a broad spectrum of established methods by executing official implementations on the specified test datasets. This approach is adopted to eliminate \textit{cross-protocol bias} that often affects academic comparisons, where performance discrepancies may arise from disparate minutiae-pairing heuristics, varying spatial or angular thresholds, or inconsistent handling of image boundaries. By processing all methods through the same evaluation pipeline described in Sec.~\ref{subsec:setup}, performance variations are attributable solely to the extraction algorithms. The selected methods represent three distinct categories:
\begin{itemize}
    \item \textit{SOTA Deep Learning Models}: FingerNet~\cite{Tang2017}, MinutiaeNet~\cite{Nguyen2018a}, and LatentAFIS~\cite{Cao2020}. These are the open-source neural architectures identified in Table~\ref{tabMethods}.
    \item \textit{COTS Solutions}: FingerJet OSE~\cite{DigitalPersona}, FDx v4.31~\cite{SecuGen}, and VeriFinger v13.1~\cite{Neurotechnology}. These engines facilitate an assessment of the industrial relevance of LEADER against production-grade software.
    \item \textit{Standard Baselines and Independent Projects}: MINDTCT~\cite{Watson2007}, FingerFlow~\cite{Arendac}, and SourceAFIS~\cite{Vazan}. This group includes traditional baselines and open-source contributions that lack peer-reviewed publications, but are accessible tools within the fingerprint community.
\end{itemize}
The comparative evaluation results, summarized in Table~\ref{tab:results_f1} and Figs.~\ref{fig:pr_curves} and \ref{fig:pr_curves_2}, indicate that LEADER consistently outperforms all competitors across all datasets and evaluation regimes. On the FVC2002 DB1-A benchmark, LEADER achieves a peak $F_1$-score of 0.92, surpassing specialized deep-learning architectures such as FingerNet (0.87) and LatentAFIS (0.84), as well as high-tier commercial engines (VeriFinger and FDx, both 0.86). Notably, LEADER demonstrates high stability under stringent threshold levels: the $F_1$-score remains as high as 0.90 at $(8\text{\,px}, \pi/10\text{\,rad})$, whereas competitors experience sharper declines. This precision is further substantiated by the PR curves in Fig.~\ref{fig:pr_curves}. The superiority of the proposed architecture is even more pronounced in the type-aware regime (Fig.~\ref{fig:pr_curves_2}), where LEADER maintains a 0.12 absolute margin over the leading commercial competitor (FDx). This suggests that the two encoder-decoder components effectively extract the high-level features necessary for accurate classification of ridge endings and bifurcations.
On the latent benchmarks, LEADER exhibits robust zero-shot generalization. On NIST SD27, it achieves the highest $F_1$-score (0.71) in the type-agnostic regime, outperforming both commercial systems (VeriFinger, 0.64) and specialized models such as FingerNet (0.67) and LatentAFIS (0.62), which were explicitly exposed to latent training data. Crucially, these results are confirmed on the large-scale NIST SD302 benchmark, where LEADER reaches an $F_1$-score of 0.69, maintaining a clear advantage over FingerNet (0.65) and VeriFinger (0.60). As shown in Fig.~\ref{fig:pr_curves_2}, the performance gap widens significantly under the type-aware evaluation regime at the most stringent threshold level: on NIST SD27, LEADER achieves an $F_1$-score of 0.47 compared to 0.35 for the second-best method (VeriFinger), representing a 34\% relative improvement; similarly, on NIST SD302, LEADER retains an $F_1$-score of 0.50 compared to 0.36 for VeriFinger, yielding a 39\% relative improvement. The statistical robustness of these findings is reinforced by the paired non-parametric bootstrap confidence intervals reported in Table~\ref{tab:results_f1}. Across most configurations, LEADER's 95\% confidence intervals do not overlap with those of competing methods, except for a minor boundary contact with the runner-up (FingerNet) on NIST SD27 ($0.71\,\pm\,0.02$ vs. $0.67\,\pm\,0.02$). Furthermore, the narrower confidence intervals observed on NIST SD302 reflect the increased statistical power afforded by the larger dataset size, confirming that LEADER's superior generalization to latent impressions is consistent and resilient to sample variance. These cross-dataset results suggest that LEADER learns domain-transferable topological features that generalize more effectively than those of larger or domain-specific models, a characteristic further analyzed in Sec.~\ref{subsec:interpretability}.

\begin{figure*}[t]
\centering
\begin{tikzpicture}
    \begin{groupplot}[
        group style={
            group size=3 by 3,
            horizontal sep=0.15cm,
            vertical sep=0.25cm,
            x descriptions at=edge bottom,
            y descriptions at=edge left,
        },
        IEEEstyle,
        width=7cm,
        height=6.5cm,
        xmin=0, xmax=1, ymin=0, ymax=1,
        grid=both,
        xtick={0,0.1,...,1},
        ytick={0,0.1,...,1},
        cycle list name=graph_precision_recall,
    ]
    \input{graph_precision_recall_I.tex}
    \end{groupplot}
    \coordinate (top_row_center) at ($(group c1r1.south)!.5!(group c3r1.south)$);
    \node[above, yshift=0.3cm,
        fill=white,
        drop shadow={
            shadow xshift=0.5ex,
            shadow yshift=-0.5ex,
            fill=black!30,
            opacity=0.4
        }, inner sep=0pt] at (top_row_center) {\pgfplotslegendfromname{single_legend_I}};
\end{tikzpicture}
\caption{PR curves for the type-agnostic regime across the three datasets (one per row). Columns represent the three threshold levels.}
\label{fig:pr_curves}
\end{figure*}

\begin{figure}[p]
    \centering
    \includegraphics[width=\linewidth]{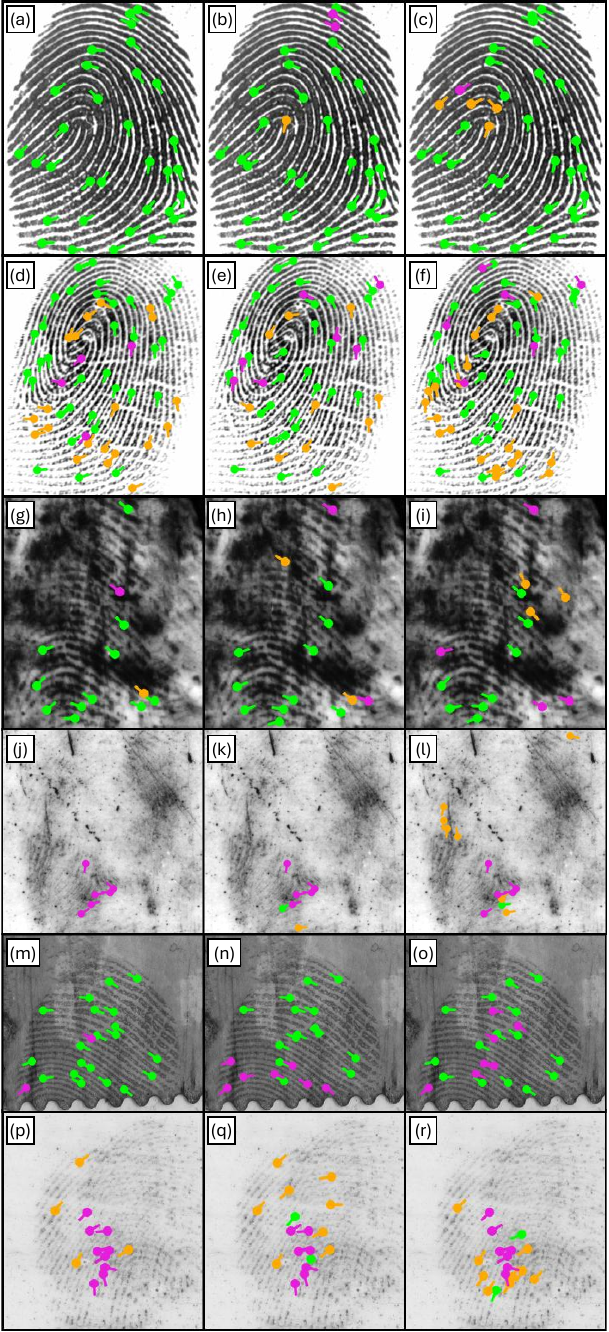}
    \caption{\textbf{Sample-level visual comparisons.} True Positives (green), False Positives (orange), and False Negatives (magenta) are overlaid for LEADER and the two top-performing competitors. 
    Rows illustrate representative performance scenarios: (a--c)~high-scoring sample on FVC2002 DB1-A; (d--f)~LEADER's lowest-performing case on FVC2002 DB1-A; (g--i)~high-scoring latent sample on NIST SD27; (j--l)~failure case on NIST SD27; (m--o)~high-scoring latent sample on NIST SD302; (p--r)~failure case on NIST SD302. First column (a, d, g, j, m, p)~shows LEADER results ($F_1$-score: 1.0, 0.75, 0.92, 0.00, 0.95, 0.00). Second and third columns display top competitors: (b)~FDx (0.95), (c)~MINDTCT~(0.93), (e)~LatentAFIS~(0.77), (f)~FingerJet~(0.71), (h)~VeriFinger~(0.83), (i)~FingerJet~(0.70), (k)~LatentAFIS~(0.25), (l)~FDx~(0.14), (n)~FingerNet~(0.81), (o)~LatentAFIS~(0.81), (q)~FingerNet~(0.21), and (r)~MINDTCT~(0.20). Full-image comparisons for all extractors are provided in Figs.~\ref{fig:good_res_plain}--\ref{fig:bad_res_latent_302} (Supplementary Material).}
    \label{fig:visual_comparison_main}
\end{figure}

To evaluate performance reliability across individual prints, the comparative analysis is extended to a sample-level ranking computed at the $(16\text{\,px}, \pi/6\text{\,rad})$ thresholds under the type-agnostic regime. Table~\ref{tab:supp_ranking} reports comprehensive ranking statistics for the three datasets: LEADER achieves the best average, top-1, and top-3 ranks and exhibits the lowest bottom half percentage on all datasets, underscoring its reliability even in challenging conditions. To provide deeper insight, Tables~\ref{tab:direct_wins_fvc}--\ref{tab:direct_wins_sd302} present pairwise direct-win matrices\footnote{These matrices also implicitly define the \textit{tie rate} between any pair of extractors, which can be calculated as the remaining percentage after accounting for their reciprocal win rates.}, where each cell $(i, j)$ denotes the percentage of samples in which method $i$ outperforms method $j$ in terms of $F_1$-score. This fine-grained comparison highlights LEADER's dominance, as it outperforms all other extractors in the vast majority of per-sample trials. On FVC2002, LEADER outperforms the runner-up (FingerNet) in 85\% of the samples, while being surpassed in only 13\% of cases (Table~\ref{tab:direct_wins_fvc}), securing first place in 71\% of all prints (Table~\ref{tab:supp_ranking}). On the latent domain, LEADER outperforms FingerNet in 63\% of the samples on both NIST SD27 and NIST SD302 (Tables~\ref{tab:direct_wins_nist}--\ref{tab:direct_wins_sd302}), remaining the top-ranked extractor in 51\% and 49\% of instances, respectively (Table~\ref{tab:supp_ranking}). This statistical consistency is qualitatively supported by the sample-level visual analysis in Fig.~\ref{fig:visual_comparison_main}. In high-performance instances (rows 1, 3, and 5), LEADER demonstrates superior spatial localization and higher detection fidelity. Even in its worst plain-fingerprint case (Fig.~\ref{fig:visual_comparison_main}d), LEADER remains competitive ($F_1$-score\,=\,0.75), with several false positives corresponding to valid minutiae omitted from the ground truth. Finally, the two failure cases in Figs.~\ref{fig:visual_comparison_main}j and~\ref{fig:visual_comparison_main}p underscore that extraction in heavily corrupted impressions remains an open challenge.

\begin{figure*}[ht]
\centering
\begin{tikzpicture}
    \begin{groupplot}[
        group style={
            group size=3 by 3,
            horizontal sep=0.15cm,
            vertical sep=0.25cm,
            x descriptions at=edge bottom,
            y descriptions at=edge left,
        },
        IEEEstyle,
        width=7cm,
        height=6.5cm,
        xmin=0, xmax=1, ymin=0, ymax=1,
        grid=both,
        xtick={0,0.1,...,1},
        ytick={0,0.1,...,1},
        cycle list name=graph_precision_recall_2,
    ]
    \input{graph_precision_recall_T.tex}
    \end{groupplot}
    \coordinate (top_row_center) at ($(group c1r1.south)!.5!(group c3r1.south)$);
    \node[above, yshift=0.3cm,
        fill=white,
        drop shadow={
            shadow xshift=0.5ex,
            shadow yshift=-0.5ex,
            fill=black!30,
            opacity=0.4
        }, inner sep=0pt] at (top_row_center) {\pgfplotslegendfromname{single_legend_T}};
\end{tikzpicture}
\caption{PR curves for the type-aware regime, adhering to the same grid layout and shared legend as in Fig.~\ref{fig:pr_curves}. Methods lacking minutiae-type support (FingerNet, LatentAFIS, and MinutiaeNet) are omitted for a fair comparison.}
\label{fig:pr_curves_2}
\end{figure*}

\begin{notablediff*}[htb]
\centering
\begin{threeparttable}
\caption{Sample-level ranking analysis. Results summarize multiple independent rankings (one per sample) based on the type-agnostic $F_1$-score at ($16\text{\,px}, \pi/6\text{\,rad}$). \textbf{Bold} and \underline{underlined} values indicate the best and second-best results for each metric, respectively.}
\label{tab:supp_ranking}
\setlength{\tabcolsep}{2.5pt} 
\begin{tabular*}{\textwidth}{@{} l crrr c crrr c crrr @{}}
\toprule
& \multicolumn{4}{c}{\textbf{FVC2002 DB1-A}} & \phantom{a} & \multicolumn{4}{c}{\textbf{NIST SD27}} & \phantom{a} & \multicolumn{4}{c}{\textbf{NIST SD302}} \\
\cmidrule(lr){2-5} \cmidrule(lr){7-10}  \cmidrule(lr){12-15}
\textbf{Method} & \textbf{Avg.\,Rank\,$\pm$\,S.D.} & \textbf{T1}\,(\%) & \textbf{T3}\,(\%) & \textbf{BH}\,(\%) && \textbf{Avg.\,Rank\,$\pm$\,S.D.} & \textbf{T1}\,(\%) & \textbf{T3}\,(\%) & \textbf{BH}\,(\%)  && \textbf{Avg.\,Rank\,$\pm$\,S.D.} & \textbf{T1}\,(\%) & \textbf{T3}\,(\%) & \textbf{BH}\,(\%)\\
\midrule

FingerNet & \underline{3.5 $\pm$ 1.9} & \underline{13}\phantom{00} & \underline{56}\phantom{00} & 19\phantom{00} & & \underline{3.0 $\pm$ 2.0} & \underline{24}\phantom{00} & \underline{68}\phantom{00} & 13\phantom{00} & & \underline{2.7 $\pm$ 1.8} & \underline{26}\phantom{00} & \underline{75}\phantom{00} & 8\phantom{00} \\
LatentAFIS & 5.0 $\pm$ 2.0 & 3\phantom{00} & 27\phantom{00} & 43\phantom{00} & & 3.7 $\pm$ 1.8 & 10\phantom{00} & 44\phantom{00} & 13\phantom{00} & & 3.4 $\pm$ 1.5 & 11\phantom{00} & 53\phantom{00} & 7\phantom{00} \\
MinutiaeNet & 9.8 $\pm$ 0.4 & 0\phantom{00} & 0\phantom{00} & 100\phantom{00} & & 5.8 $\pm$ 2.3 & 3\phantom{00} & 16\phantom{00} & 54\phantom{00} & & 6.8 $\pm$ 2.1 & 1\phantom{00} & 6\phantom{00} & 70\phantom{00} \\
\addlinespace[0.25em]
FingerJet & 5.8 $\pm$ 1.7 & 1\phantom{00} & 10\phantom{00} & 62\phantom{00} & & 7.6 $\pm$ 1.9 & 0\phantom{00} & 2\phantom{00} & 84\phantom{00} & & 7.6 $\pm$ 2.2 & 1\phantom{00} & 5\phantom{00} & 83\phantom{00} \\
FDx & 4.0 $\pm$ 1.9 & 8\phantom{00} & 43\phantom{00} & 22\phantom{00} & & 6.2 $\pm$ 1.8 & 1\phantom{00} & 7\phantom{00} & 60\phantom{00} & & 5.6 $\pm$ 1.9 & 3\phantom{00} & 13\phantom{00} & 52\phantom{00} \\
VeriFinger & 3.7 $\pm$ 1.5 & 5\phantom{00} & 46\phantom{00} & \underline{12}\phantom{00} & & 3.2 $\pm$ 1.4 & 12\phantom{00} & 65\phantom{00} & \underline{10}\phantom{00} & & 3.0 $\pm$ 1.5 & 17\phantom{00} & 65\phantom{00} & \underline{6}\phantom{00} \\
\addlinespace[0.25em]
MINDTCT & 5.1 $\pm$ 1.9 & 4\phantom{00} & 25\phantom{00} & 47\phantom{00} & & 8.6 $\pm$ 1.4 & 0\phantom{00} & 0\phantom{00} & 95\phantom{00} & & 6.8 $\pm$ 1.8 & 1\phantom{00} & 6\phantom{00} & 81\phantom{00} \\
FingerFlow & 9.2 $\pm$ 0.5 & 0\phantom{00} & 0\phantom{00} & 100\phantom{00} & & 5.8 $\pm$ 2.3 & 7\phantom{00} & 18\phantom{00} & 62\phantom{00} & & 6.9 $\pm$ 2.0 & 1\phantom{00} & 5\phantom{00} & 76\phantom{00} \\
SourceAFIS & 7.1 $\pm$ 1.3 & 0\phantom{00} & 2\phantom{00} & 86\phantom{00} & & 8.4 $\pm$ 1.3 & 0\phantom{00} & 0\phantom{00} & 99\phantom{00} & & 6.6 $\pm$ 1.8 & 1\phantom{00} & 5\phantom{00} & 77\phantom{00} \\
\addlinespace[0.25em]
LEADER (Ours) & \textbf{1.4 $\pm$ 0.8} & \textbf{71}\phantom{00} & \textbf{96}\phantom{00} & \textbf{0}\phantom{00} & & \textbf{2.0 $\pm$ 1.3} & \textbf{51}\phantom{00} & \textbf{87}\phantom{00} & \textbf{1}\phantom{00} & & \textbf{1.9 $\pm$ 1.2} & \textbf{49}\phantom{00} & \textbf{91}\phantom{00} & \textbf{1}\phantom{00} \\

\bottomrule
\end{tabular*}
\begin{tablenotes}
\item \textbf{Avg.\,Rank\,$\pm$\,S.D.}: Mean rank and standard deviation across all samples (lower is better).
\item \textbf{T1}, \textbf{T3}: Percentage of samples achieving a Top-1 or Top-3 rank, respectively.
\item \textbf{BH} (Bottom Half): Percentage of samples falling into the lower 50\% of the ranking (lower is better).
\end{tablenotes}
\end{threeparttable}
\end{notablediff*}

\begin{notablediff}[ht]
\centering
\scriptsize
\renewcommand{\arraystretch}{0.8}
\begin{threeparttable}
\caption{Pairwise direct-win rates (\%) on FVC2002 DB1-A based on type-agnostic $F_1$-score at ($16\text{\,px}, \pi/6\text{\,rad}$). Cell $(i,j)$ indicates row $i$ outperforming column $j$. \textbf{Bold} values denote LEADER's direct-win rates.}
\label{tab:direct_wins_fvc}
\setlength{\tabcolsep}{3pt} 
\begin{tabular}{@{} l *{10}{r} @{}}
\toprule
& \rotatebox{90}{FingerNet} & \rotatebox{90}{LatentAFIS} & \rotatebox{90}{MinutiaeNet} & \rotatebox{90}{FingerJet} & \rotatebox{90}{FDx} & \rotatebox{90}{VeriFinger} & \rotatebox{90}{MINDTCT} & \rotatebox{90}{FingerFlow} & \rotatebox{90}{SourceAFIS} & \rotatebox{90}{LEADER} \\
\midrule
FingerNet & --- &  71 & 100 &  81 &  63 &  55 &  72 & 100 &  91 &  13 \\
LatentAFIS &  28 & --- & 100 &  59 &  34 &  30 &  54 &  99 &  82 &   5 \\
MinutiaeNet &   0 &   0 & --- &   1 &   0 &   0 &   0 &  19 &   0 &   0 \\
FingerJet &  19 &  40 &  99 & --- &  26 &  17 &  37 &  99 &  72 &   2 \\
FDx &  36 &  63 & 100 &  73 & --- &  44 &  66 & 100 &  93 &   9 \\
VeriFinger &  44 &  68 & 100 &  80 &  51 & --- &  69 & 100 &  92 &   7 \\
MINDTCT &  27 &  44 & 100 &  60 &  34 &  31 & --- & 100 &  83 &   6 \\
FingerFlow &   0 &   1 &  80 &   1 &   0 &   0 &   0 & --- &   1 &   0 \\
SourceAFIS &   9 &  17 & 100 &  27 &   7 &   7 &  15 &  99 & --- &   1 \\
LEADER (Ours) & \textbf{ 85} & \textbf{ 94} & \textbf{100} & \textbf{ 98} & \textbf{ 89} & \textbf{ 90} & \textbf{ 93} & \textbf{100} & \textbf{ 99} & \textbf{---} \\
\bottomrule
\end{tabular}
\end{threeparttable}
\end{notablediff}

\begin{notablediff}[ht]
\centering
\scriptsize
\renewcommand{\arraystretch}{0.8}
\begin{threeparttable}
\caption{Pairwise direct-win rates (\%) on NIST SD27. Notation as in Table~\ref{tab:direct_wins_fvc}.}
\label{tab:direct_wins_nist}
\setlength{\tabcolsep}{3pt} 
\begin{tabular}{@{} l *{10}{r} @{}}
\toprule
& \rotatebox{90}{FingerNet} & \rotatebox{90}{LatentAFIS} & \rotatebox{90}{MinutiaeNet} & \rotatebox{90}{FingerJet} & \rotatebox{90}{FDx} & \rotatebox{90}{VeriFinger} & \rotatebox{90}{MINDTCT} & \rotatebox{90}{FingerFlow} & \rotatebox{90}{SourceAFIS} & \rotatebox{90}{LEADER} \\
\midrule
FingerNet & --- &  63 &  79 &  94 &  86 &  55 &  96 &  79 &  96 &  32 \\
LatentAFIS &  33 & --- &  75 &  92 &  85 &  39 &  95 &  76 &  97 &  18 \\
MinutiaeNet &  19 &  21 & --- &  69 &  55 &  21 &  82 &  39 &  82 &  10 \\
FingerJet &   5 &   6 &  31 & --- &  31 &   3 &  58 &  32 &  58 &   1 \\
FDx &  13 &  13 &  44 &  68 & --- &   7 &  87 &  47 &  87 &   2 \\
VeriFinger &  45 &  60 &  79 &  97 &  92 & --- & 100 &  80 & 100 &  24 \\
MINDTCT &   2 &   3 &  17 &  31 &   9 &   0 & --- &  17 &  38 &   1 \\
FingerFlow &  20 &  23 &  49 &  68 &  52 &  19 &  82 & --- &  80 &  10 \\
SourceAFIS &   3 &   3 &  18 &  40 &  13 &   0 &  61 &  19 & --- &   0 \\
LEADER (Ours) & \textbf{ 63} & \textbf{ 80} & \textbf{ 88} & \textbf{ 99} & \textbf{ 98} & \textbf{ 73} & \textbf{ 99} & \textbf{ 89} & \textbf{100} & \textbf{---} \\
\bottomrule
\end{tabular}
\end{threeparttable}
\end{notablediff}

\begin{notablediff}[ht]
\centering
\scriptsize
\renewcommand{\arraystretch}{0.8}
\begin{threeparttable}
\caption{Pairwise direct-win rates (\%) on NIST SD302. Notation as in Table~\ref{tab:direct_wins_fvc}.}
\label{tab:direct_wins_sd302}
\setlength{\tabcolsep}{3pt} 
\begin{tabular}{@{} l *{10}{r} @{}}
\toprule
& \rotatebox{90}{FingerNet} & \rotatebox{90}{LatentAFIS} & \rotatebox{90}{MinutiaeNet} & \rotatebox{90}{FingerJet} & \rotatebox{90}{FDx} & \rotatebox{90}{VeriFinger} & \rotatebox{90}{MINDTCT} & \rotatebox{90}{FingerFlow} & \rotatebox{90}{SourceAFIS} & \rotatebox{90}{LEADER} \\
\midrule
FingerNet & --- &  63 &  87 &  89 &  83 &  55 &  87 &  88 &  88 &  32 \\
LatentAFIS &  32 & --- &  93 &  92 &  82 &  40 &  91 &  93 &  92 &  18 \\
MinutiaeNet &   8 &   5 & --- &  66 &  36 &   6 &  51 &  44 &  44 &   3 \\
FingerJet &   3 &   5 &  25 & --- &  12 &   4 &  18 &  25 &  23 &   2 \\
FDx &  11 &  15 &  57 &  59 & --- &  11 &  57 &  58 &  56 &   5 \\
VeriFinger &  39 &  55 &  90 &  91 &  84 & --- &  90 &  90 &  92 &  24 \\
MINDTCT &   5 &   6 &  40 &  45 &  15 &   5 & --- &  42 &  33 &   2 \\
FingerFlow &   6 &   4 &  40 &  62 &  33 &   5 &  47 & --- &  40 &   2 \\
SourceAFIS &   7 &   6 &  50 &  57 &  26 &   5 &  47 &  53 & --- &   2 \\
LEADER (Ours) & \textbf{ 63} & \textbf{ 78} & \textbf{ 95} & \textbf{ 96} & \textbf{ 93} & \textbf{ 72} & \textbf{ 96} & \textbf{ 96} & \textbf{ 97} & \textbf{---} \\
\bottomrule
\end{tabular}
\end{threeparttable}
\end{notablediff}

Regarding computational efficiency, Table~\ref{tab:runtime} reports average processing times. With a compact footprint of 0.9\,M parameters, the proposed architecture achieves an average GPU inference time of 15\,ms. LEADER maintains a competitive runtime of 322\,ms on CPU, outperforming commercial engines such as FDx and VeriFinger. While the CPU execution of competing deep-learning architectures~\cite{Tang2017, Cao2020, Nguyen2018a} scales poorly, the proposed model remains viable for real-time applications on general-purpose processors. Modern community-driven implementations such as FingerFlow, which extend~\cite{Nguyen2018a} with additional neural stages, exhibit significantly higher latency (7068\,ms on GPU), justifying the lightweight and integrated design of LEADER. These results indicate a favorable trade-off between structural complexity and throughput. Beyond inference speed, an analytical memory footprint estimation was conducted. For single-image inference at the average image size reported in Table~\ref{tab:runtime}, accounting for model parameters and peak tensor working set—including intermediate activation maps and skip-connection buffers retained across cascaded blocks—the required memory remains under 20\,MB with 32-bit single-precision floating-point representation. This minimal runtime memory overhead confirms that the architecture is well-suited for both high-throughput server deployments and memory-constrained edge processing.

\begin{table}[t]
\centering
\begin{threeparttable}
\caption{Average execution time (ms) on NIST SD27\tnote{1}.}
\label{tab:runtime}
\begin{tabular*}{\columnwidth}{@{} l l @{\extracolsep{\fill}} S[table-format=3.1] S[table-format=5.0] @{}}
\toprule
& \textbf{Method} & {\textbf{Parameters (M)}} & {\textbf{Time (ms)}} \\
\midrule
\multirow{10}{*}{\rotatebox[origin=c]{90}{\textsc{cpu}\tnote{2}}} 
& FingerNet & 4.7 & 3964 \\
& LatentAFIS & 5.5 & 5860 \\
& MinutiaeNet & 74.4 & 25744 \\
\addlinespace[0.25em]
& FingerJet & {---} & 63\rlap{\tnote{3}} \\
& FDx & {---} & 397 \\
& VeriFinger & {---} & 1199 \\
\addlinespace[0.25em]
& MINDTCT & {---} & 177\rlap{\tnote{3}} \\
& FingerFlow & 291.3 & 15110 \\
& SourceAFIS & {---} & 101 \\
\addlinespace[0.25em]
& LEADER (Ours) & \bfseries 0.9 & \bfseries 322 \\
\midrule
\multirow{5}{*}{\rotatebox[origin=c]{90}{\textsc{gpu}\phantom{\tnote{1}}}}
& FingerNet & 4.7 & 600\rlap{\tnote{4}} \\
& LatentAFIS & 5.5 & 1500\rlap{\tnote{4}} \\
& MinutiaeNet & 74.4 & 1500\rlap{\tnote{4}} \\
\addlinespace[0.25em]
& FingerFlow & 291.3 & 7068\rlap{\tnote{5}} \\
\addlinespace[0.25em]
& LEADER (Ours) & \bfseries 0.9 & \bfseries 15\rlap{\normalfont\tnote{5,6}} \\
\bottomrule
\end{tabular*}
\begin{tablenotes}
\item[1] Average image dimensions: $315 \times 397$ pixels.
\item[2] Intel\textregistered\ Xeon\textregistered\ Silver 4112 CPU @ 2.60\,GHz. All methods evaluated on CPU were executed with a batch size of one.
\item[3] Standalone executable invocation; includes minimal disk I/O overhead ($\sim$1\,ms). All other methods evaluate in-memory execution excluding disk I/O, but strictly including all preprocessing and postprocessing stages.
\item[4] GPU runtimes reported from original papers~\cite{Tang2017, Cao2020, Nguyen2018a} due to legacy framework dependencies incompatible with modern software stacks.
\item[5] NVIDIA RTX™ 3080 Ti GPU.
\item[6] Average inference time measured with batch size 32.
\end{tablenotes}
\end{threeparttable}
\end{table}

\subsection{Ablation Study}
\label{subsec:ablation}
To evaluate the contribution of individual architectural components and ground-truth encoding strategies, Table~\ref{tab:ablation_full} summarizes a comprehensive ablation study across the three test datasets. 

\subsubsection{Comparison with Iso-Parametric Baseline}
Row~\#1 evaluates an iso-parametric control model (standard encoder-decoder network with skip connections and 0.9\,M parameters, trained under identical CMR supervision). While obtaining reasonable results on plain fingerprints ($F_1$-score\,=\,0.90 vs. 0.92 for LEADER, with gaps up to 0.06 under tighter thresholds and type-aware evaluation), this baseline degrades sharply on latents (e.g., $F_1$-score of 0.55 on NIST SD27 and 0.56 on NIST SD302, vs. 0.71 and 0.69 for LEADER). These substantial performance gaps (0.16 and 0.13, respectively) demonstrate that standard feature integration is insufficient for degraded ridge patterns, confirming the structural necessity of LEADER's multi-stage pipeline.

\subsubsection{Impact of Ground-Truth Encoding}
Comparing CMR ($\boxplus$, row \#0) against Concentric Zones (\bullseye, row \#2)~\cite{Feng2023} and Gaussian Masks ($\bullet$, row \#3) highlights the role of ground-truth encoding, following a progression from more sophisticated to simpler supervision strategies. At the first threshold level, the type-agnostic $F_1$-score on NIST SD27 drops from 0.71 (CMR) to 0.63 (Concentric Zones) and 0.56 (Gaussian Masks), with a parallel decline on NIST SD302 (0.69\,$>$\,0.65\,$>$\,0.59). While Concentric Zones improve jitter tolerance over standard Gaussian masks by introducing a neutral buffer, both rely on fixed targets prone to merging in dense clusters. Conversely, CMR prevents target fusion via its adaptive morphology and enforces sharp localization through its penalty wall. Although strict geometric decoupling of each sub-element is constrained by their joint formulation, these two variants isolate the incremental benefits of the uncertainty buffer (Gaussian Masks vs. Concentric Zones) and the penalty wall with core adaptation (Concentric Zones vs. CMR), thus validating the CMR's design principles.

\subsubsection{Architectural Component Suppressions}
Single-component ablations (rows \#4--\#7) show that latent extraction relies heavily on the complete pipeline. At the first threshold level, removing the \textit{Stem} (\#4) or \textit{Context-Enc-Dec} (\#5) reduces NIST SD27 type-agnostic $F_1$-score to 0.66 and 0.64, respectively. The \textit{Refinement-Enc-Dec} (\#7) is the most critical component, whose omission drops performance to 0.59 on NIST SD27 and 0.60 on NIST SD302.

\subsubsection{Synergistic Interactions and Degraded Regimes}
Combined ablations (rows \#8--\#11) reveal non-linear coupling between representation and supervision. Jointly removing the \textit{Stem} and switching to Gaussian Masks (\#8) triggers a drop of 0.33 on NIST SD27 (first threshold level, type-agnostic $F_1$-score\,=\,0.38), whereas single suppressions reduce performance by 0.05 (\#4) and 0.15 (\#3), respectively. Finally, stripping all components down to the bare \textit{Head} (\#11) leads to complete generalization failure ($F_1$-score\,=\,0.00).

\begin{notablediff*}[htb]
\centering
\begin{threeparttable}
\caption{Ablation Study. Performance ($F_1$-score) evaluated across the three threshold levels for both type-agnostic and type-aware (in parentheses) regimes. LEADER (\#0) is evaluated against an iso-parametric baseline, two encoding variants, single-component suppressions, and progressive architectural degradations across the three datasets.}
    \label{tab:ablation_full}
    \footnotesize
    \setlength{\tabcolsep}{3.2pt}
    \begin{tabular}{cccccc ccc ccc ccc}
    \toprule
    \multicolumn{6}{c}{\textbf{Configuration}\tnote{1}} & \multicolumn{3}{c}{\textbf{FVC2002 DB1-A}} & \multicolumn{3}{c}{\textbf{NIST SD27}}  & \multicolumn{3}{c}{\textbf{NIST SD302}} \\
    \cmidrule(lr){1-6} \cmidrule(lr){7-9} \cmidrule(lr){10-12} \cmidrule(lr){13-15}
    \textbf{\#} & \textbf{S} & \textbf{CED} & \textbf{AG} & \textbf{RED} & \textbf{GTE}
    & \thrl{16}{6} & \thrl{12}{8} & \thrl{8}{10}
    & \thrl{16}{6} & \thrl{12}{8} & \thrl{8}{10}
    & \thrl{16}{6} & \thrl{12}{8} & \thrl{8}{10}\\
    \midrule

    0 & $\checkmark$ & $\checkmark$ & $\checkmark$ & $\checkmark$ & $\boxplus$ & $\mathbf{0.92\,\text{\scriptsize(0.85)}}$ & $\mathbf{0.91\,\text{\scriptsize(0.84)}}$ & $\mathbf{0.90\,\text{\scriptsize(0.83)}}$ & $\mathbf{0.71\,\text{\scriptsize(0.52)}}$ & $\mathbf{0.69\,\text{\scriptsize(0.51)}}$ & $\mathbf{0.62\,\text{\scriptsize(0.47)}}$ & $\mathbf{0.69\,\text{\scriptsize(0.53)}}$ & $\mathbf{0.67\,\text{\scriptsize(0.52)}}$ & $\mathbf{0.64\,\text{\scriptsize(0.50)}}$ \\
    \addlinespace[0.25em]
    1 & $\ast$ & $\ast$ & $\ast$ & $\ast$ & $\boxplus$ & $0.90\,\text{\scriptsize(0.80)}$ & $0.89\,\text{\scriptsize(0.79)}$ & $0.86\,\text{\scriptsize(0.77)}$ & $0.55\,\text{\scriptsize(0.39)}$ & $0.50\,\text{\scriptsize(0.36)}$ & $0.44\,\text{\scriptsize(0.32)}$ & $0.56\,\text{\scriptsize(0.42)}$ & $0.54\,\text{\scriptsize(0.40)}$ & $0.50\,\text{\scriptsize(0.38)}$ \\
    \addlinespace[0.25em]
    2 & $\checkmark$ & $\checkmark$ & $\checkmark$ & $\checkmark$ & \bullseye & $0.91\,\text{\scriptsize(0.84)}$ & $0.90\,\text{\scriptsize(0.83)}$ & $0.89\,\text{\scriptsize(0.82)}$ & $0.63\,\text{\scriptsize(0.47)}$ & $0.60\,\text{\scriptsize(0.45)}$ & $0.54\,\text{\scriptsize(0.41)}$ & $0.65\,\text{\scriptsize(0.51)}$ & $0.64\,\text{\scriptsize(0.50)}$ & $0.61\,\text{\scriptsize(0.48)}$ \\
    3 & $\checkmark$ & $\checkmark$ & $\checkmark$ & $\checkmark$ & $\bullet$ & $0.90\,\text{\scriptsize(0.83)}$ & $0.90\,\text{\scriptsize(0.82)}$ & $0.88\,\text{\scriptsize(0.81)}$ & $0.56\,\text{\scriptsize(0.43)}$ & $0.53\,\text{\scriptsize(0.41)}$ & $0.47\,\text{\scriptsize(0.37)}$ & $0.59\,\text{\scriptsize(0.48)}$ & $0.58\,\text{\scriptsize(0.46)}$ & $0.55\,\text{\scriptsize(0.45)}$ \\
    \addlinespace[0.25em]
    4 & -- & $\checkmark$ & $\checkmark$ & $\checkmark$ & $\boxplus$ & $0.91\,\text{\scriptsize(0.82)}$ & $0.90\,\text{\scriptsize(0.81)}$ & $0.88\,\text{\scriptsize(0.80)}$ & $0.66\,\text{\scriptsize(0.47)}$ & $0.63\,\text{\scriptsize(0.45)}$ & $0.56\,\text{\scriptsize(0.41)}$ & $0.64\,\text{\scriptsize(0.47)}$ & $0.62\,\text{\scriptsize(0.46)}$ & $0.58\,\text{\scriptsize(0.43)}$ \\
    5 & $\checkmark$ & -- & $\checkmark$ & $\checkmark$ & $\boxplus$ & $0.91\,\text{\scriptsize(0.83)}$ & $0.91\,\text{\scriptsize(0.83)}$ & $0.90\,\text{\scriptsize(0.82)}$ & $0.64\,\text{\scriptsize(0.46)}$ & $0.61\,\text{\scriptsize(0.44)}$ & $0.54\,\text{\scriptsize(0.41)}$ & $0.64\,\text{\scriptsize(0.48)}$ & $0.62\,\text{\scriptsize(0.46)}$ & $0.59\,\text{\scriptsize(0.45)}$ \\
    6 & $\checkmark$ & $\checkmark$ & -- & $\checkmark$ & $\boxplus$ & $0.91\,\text{\scriptsize(0.84)}$ & $0.91\,\text{\scriptsize(0.83)}$ & $0.90\,\text{\scriptsize(0.82)}$ & $0.64\,\text{\scriptsize(0.46)}$ & $0.61\,\text{\scriptsize(0.45)}$ & $0.55\,\text{\scriptsize(0.41)}$ & $0.64\,\text{\scriptsize(0.49)}$ & $0.62\,\text{\scriptsize(0.47)}$ & $0.59\,\text{\scriptsize(0.45)}$ \\
    7 & $\checkmark$ & $\checkmark$ & $\checkmark$ & -- & $\boxplus$ & $0.90\,\text{\scriptsize(0.81)}$ & $0.89\,\text{\scriptsize(0.80)}$ & $0.87\,\text{\scriptsize(0.79)}$ & $0.59\,\text{\scriptsize(0.43)}$ & $0.56\,\text{\scriptsize(0.41)}$ & $0.49\,\text{\scriptsize(0.37)}$ & $0.60\,\text{\scriptsize(0.46)}$ & $0.58\,\text{\scriptsize(0.45)}$ & $0.55\,\text{\scriptsize(0.43)}$ \\
    \addlinespace[0.25em]
    8 & -- & $\checkmark$ & $\checkmark$ & $\checkmark$ & $\bullet$ & $0.85\,\text{\scriptsize(0.74)}$ & $0.82\,\text{\scriptsize(0.71)}$ & $0.76\,\text{\scriptsize(0.67)}$ & $0.38\,\text{\scriptsize(0.28)}$ & $0.33\,\text{\scriptsize(0.25)}$ & $0.27\,\text{\scriptsize(0.21)}$ & $0.41\,\text{\scriptsize(0.32)}$ & $0.37\,\text{\scriptsize(0.29)}$ & $0.32\,\text{\scriptsize(0.25)}$ \\
    9 & -- & -- & $\checkmark$ & $\checkmark$ & $\bullet$ & $0.71\,\text{\scriptsize(0.60)}$ & $0.65\,\text{\scriptsize(0.56)}$ & $0.58\,\text{\scriptsize(0.51)}$ & $0.28\,\text{\scriptsize(0.20)}$ & $0.22\,\text{\scriptsize(0.16)}$ & $0.16\,\text{\scriptsize(0.12)}$ & $0.26\,\text{\scriptsize(0.20)}$ & $0.22\,\text{\scriptsize(0.16)}$ & $0.17\,\text{\scriptsize(0.14)}$ \\
    10 & -- & -- & -- & $\checkmark$ & $\bullet$ & $0.70\,\text{\scriptsize(0.59)}$ & $0.64\,\text{\scriptsize(0.55)}$ & $0.56\,\text{\scriptsize(0.49)}$ & $0.26\,\text{\scriptsize(0.19)}$ & $0.21\,\text{\scriptsize(0.15)}$ & $0.15\,\text{\scriptsize(0.12)}$ & $0.24\,\text{\scriptsize(0.18)}$ & $0.20\,\text{\scriptsize(0.15)}$ & $0.16\,\text{\scriptsize(0.12)}$ \\
    11 & -- & -- & -- & -- & $\bullet$ & $0.00\,\text{\scriptsize(0.00)}$ & $0.00\,\text{\scriptsize(0.00)}$ & $0.00\,\text{\scriptsize(0.00)}$ & $0.00\,\text{\scriptsize(0.00)}$ & $0.00\,\text{\scriptsize(0.00)}$ & $0.00\,\text{\scriptsize(0.00)}$ & $0.00\,\text{\scriptsize(0.00)}$ & $0.00\,\text{\scriptsize(0.00)}$ & $0.00\,\text{\scriptsize(0.00)}$ \\

    \bottomrule
    \end{tabular}
    \begin{tablenotes}
        \item[1] \textbf{Architectural components} (\textbf{S}: Stem; \textbf{CED}: Context-Enc-Dec; \textbf{AG}: Attention Gate; \textbf{RED}: Refinement-Enc-Dec): $\checkmark$~Component adopted; \mbox{--~Component} suppressed; $\ast$\;$\ast$\;$\ast$\;$\ast$~Entire architecture replaced by an iso-parametric baseline (standard encoder-decoder with $0.9\,\text{M}$ parameters). \\
        \textbf{Ground-Truth Encoding (GTE)}: $\boxplus$ CMR, \bullseye Concentric Zones~\cite{Feng2023}, $\bullet$ Gaussian Masks.
    \end{tablenotes}
\end{threeparttable}
\end{notablediff*}

\subsection{Visual Analysis and Feature Interpretability}
\label{subsec:interpretability}

\begin{figure}[b]
\centering
\includegraphics[width=\linewidth]{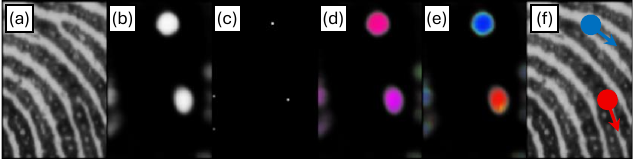}
\caption{\textbf{Visualization of output maps and decoding logic.} (a) Input fingerprint region. (b) Predicted position map $\widehat{\mathbf{P}}$. (c) Refined map $\widetilde{\mathbf{P}}$ showing local peaks obtained via on-graph NMS. (d) Direction map $\widehat{\mathbf{D}}$: the angle $\hat{d}_{ij}$ is mapped to the \textit{Hue} component (HSV); the two blobs show distinct shades of magenta, corresponding to the directions of the two minutiae, approximately $-\pi/5$ (top) and $-\pi/3$ (bottom). (e) Type map $\widehat{\mathbf{T}}$: \textit{Hue} encodes the classification score $\hat{t}_{ij}$ on a gradient from red (bifurcation) to blue (ridge ending). In (d) and (e), \textit{Saturation} and \textit{Value} are modulated by $\hat{p}_{ij}$ to suppress background noise. (f) Final extracted set $\widehat{\mathbb{M}}$.}
\label{fig:in_out}
\end{figure}

Fig.~\ref{fig:in_out} visualizes the output maps generated for a high-quality fingerprint region. The position map $\widehat{\mathbf{P}}$ (Fig.~\ref{fig:in_out}b) and its refined version $\widetilde{\mathbf{P}}$ (Fig.~\ref{fig:in_out}c) demonstrate how the network translates ridge topology into localized confidence peaks. Notably, only genuine minutiae survive the threshold $\tau_q=0.6$, an operating point well suited for high-quality plain fingerprint extraction. The direction and type maps, $\widehat{\mathbf{D}}$ (Fig.~\ref{fig:in_out}d) and $\widehat{\mathbf{T}}$ (Fig.~\ref{fig:in_out}e), confirm the precision of the \textit{Head} in regressing directions and discriminating types, resulting in a highly accurate set $\widehat{\mathbb{M}}$.

\begin{figure*}[ht]
    \centering
    \includegraphics[width=\textwidth]{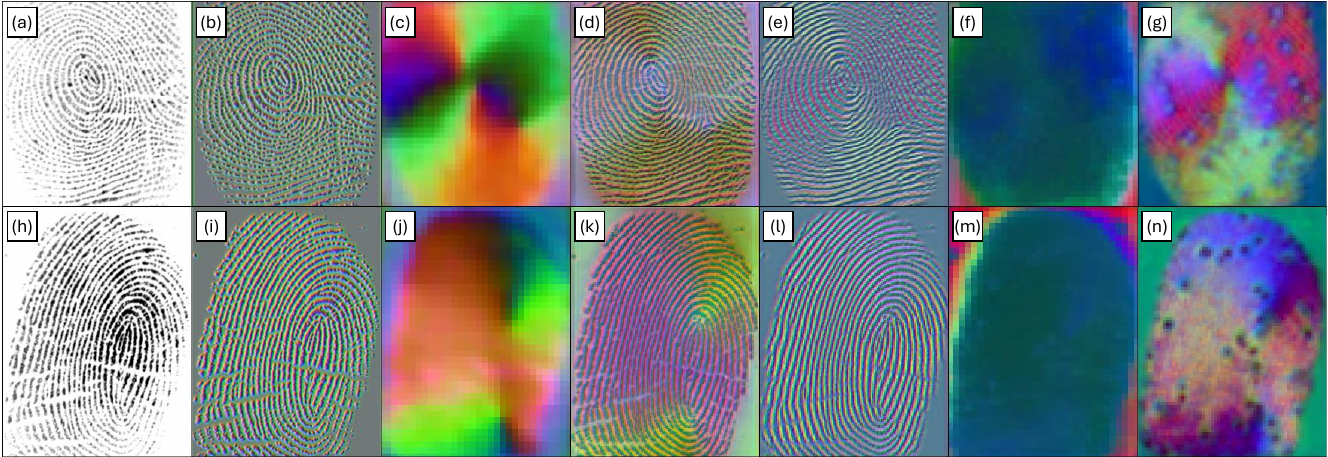}
    \caption{\textbf{Visual analysis of internal features through PCA projection.} (a, h) Input fingerprints from FVC2002 DB1-A. (b--g, i--n) RGB visualization of the first three principal components of the feature maps. All visualized activations are generated via GELU, except for (d, k), which correspond to the Sigmoid gating signal $\mathbf{X}_{gate}$ from the \textit{Attention-Gate}. (b, i) First \textit{StemBlock}. (c, j) Second \textit{SeparableConvBlock} in the encoding path of the \textit{Context-Enc-Dec}. The remaining panels illustrate successive \textit{InvBottleneckConvBlock} activations within the \textit{Refinement-Enc-Dec}: (e, l) first in encoding, (f, m) penultimate in decoding, and (g, n) final in decoding.}
    \label{fig:features_pca}
\end{figure*}

\begin{figure}[t]
    \centering
    \includegraphics[width=\linewidth]{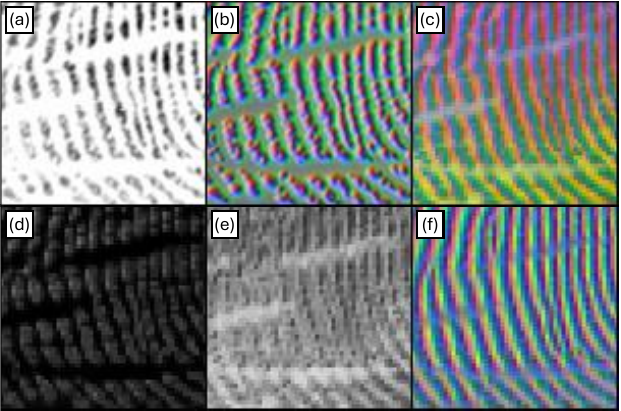}
    \caption{\textbf{Analysis of ridge cut robustness.} The sequence focuses on the bottom-left region of the fingerprint in Fig.~\ref{fig:features_pca}h: (a) raw input; (b) PCA projection of \textit{Stem} features; (c) PCA projection of the \textit{Attention-Gate} signal $\mathbf{X}_{gate}$; (d, e) individual $\mathbf{X}_{gate}$ channels exhibiting noise reduction and localized noise focusing; (f) topological inpainting in the PCA projection of \textit{Refinement-Enc-Dec} features.}
    \label{fig:scratches}
\end{figure}

\begin{figure}[b]
    \centering
    \includegraphics[width=\linewidth]{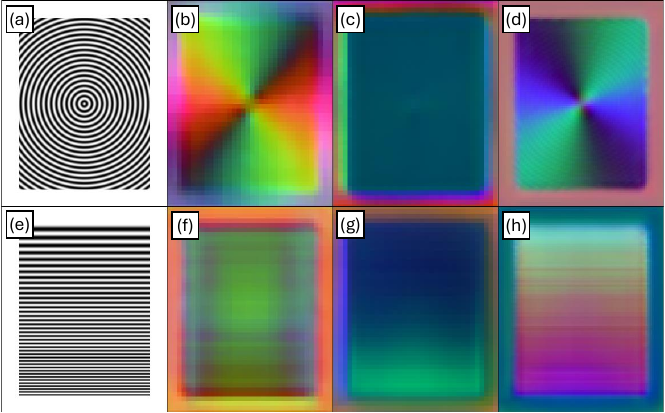}
    \caption{\textbf{Internal representation of synthetic geometric patterns.} Features are visualized via PCA projection. Top row (a--d): Concentric circles (constant frequency). Bottom row (e--h): Horizontal ridges (varying frequency). Columns: (a, e) input; (b, f) orientation-sensitive features (\textit{Context-Enc-Dec}); (c, g) frequency/segmentation features (\textit{Refinement-Enc-Dec}); (d, h) final common representation.}
    \label{fig:synthetic}
\end{figure}

While the output maps in Fig.~\ref{fig:in_out} demonstrate decoding accuracy, they do not reveal how the architecture captures fingerprint topology. As a qualitative visual diagnostic, Principal Component Analysis (PCA) is applied to the activations of key layers. For each selected layer, feature vectors are collected from all pixels across the first half of the FVC2002 DB1-A dataset. These high-dimensional vectors are projected onto a three-dimensional subspace defined by the first three principal components and normalized within the global range. Mapping these components to the RGB color space enables direct visualization of the internal representation, as illustrated in Fig.~\ref{fig:features_pca}.
Early representations produced by the \textit{Stem} (Figs.~\ref{fig:features_pca}b and~\ref{fig:features_pca}i) emphasize local gradients, serving as a multi-scale edge detection stage. A more complex behavior emerges in the \textit{Context-Enc-Dec} (Figs.~\ref{fig:features_pca}c and~\ref{fig:features_pca}j), where the color distribution correlates with local ridge orientation. Notably, the network has autonomously discovered the importance of the orientation field---a cornerstone of traditional fingerprint analysis~\cite{Cappelli2024}---and learned to encode it without explicit supervision (as validated by the ideal angular response in Fig.~\ref{fig:synthetic}b).
The \textit{Attention-Gate} signal $\mathbf{X}_{gate}$ (Figs.~\ref{fig:features_pca}d and~\ref{fig:features_pca}k) reveals a dual role: it tracks both orientation and localized noise, such as the prominent ridge cuts in the second sample, which are selectively dampened to stabilize subsequent extraction (see Fig.~\ref{fig:scratches}). In the early encoding stages of the \textit{Refinement-Enc-Dec} (Figs.~\ref{fig:features_pca}e and~\ref{fig:features_pca}l), the model performs a \textit{neural thinning} process. The network generates an enhanced skeleton-like representation of ridges and valleys, effectively using a learned inpainting mechanism to close gaps caused by cuts. This suggests the emergence of a topological prior that maintains ridge continuity (see Fig.~\ref{fig:scratches}f).
In the final stages of the \textit{Refinement-Enc-Dec} (Figs.~\ref{fig:features_pca}f and~\ref{fig:features_pca}m), the dominant feature becomes foreground-background segmentation. While traditional methods typically perform segmentation as a preprocessing step~\cite{Cappelli2023}, LEADER treats it as a high-level semantic attribute learned during final refinement. This layer also appears to encode local ridge frequency~\cite{Cappelli2024a}, whose emergence is isolated in the frequency-gradient analysis of Fig.~\ref{fig:synthetic}g. Finally, the pre-head feature maps (Figs.~\ref{fig:features_pca}g and~\ref{fig:features_pca}n) consolidate all relevant information, effectively encoding both orientation and minutiae candidates to provide the necessary context for task-specific processing.

To investigate LEADER's robustness, Fig.~\ref{fig:scratches} illustrates its response to deep ridge cuts. While early \textit{Stem} activations (Fig.~\ref{fig:scratches}b) treat these scratches as valid structural gradients, the \textit{Attention-Gate} performs a multi-channel noise decomposition. Individual channels of $\mathbf{X}_{gate}$ exhibit specialized behaviors: some serve as suppression masks by mapping noise to near-zero values (Fig.~\ref{fig:scratches}d), while others highlight noise patterns (Fig.~\ref{fig:scratches}e), likely providing a spatial \textit{invalidity map} to guide subsequent reconstruction. The resulting neural thinning (Fig.~\ref{fig:scratches}f) appears to visually compensate for local interruptions, exhibiting a behavior resembling topological inpainting. Ridges are reconnected across the cuts with high chromatic continuity; the slightly fainter hue observed at scratch locations suggests that the network projects the ridge flow based on global context while retaining a trace of the missing original information. This emergent ability to restore ridge topology represents a significant advantage over traditional processing pipelines.

To isolate the response to fundamental geometric properties, LEADER was evaluated on two synthetic patterns: concentric circles with constant frequency (Fig.~\ref{fig:synthetic}a) and horizontal ridges with varying frequency (Fig.~\ref{fig:synthetic}e). PCA projections confirm the decoupling of these emergent features. In the \textit{Context-Enc-Dec}, the model behaves as a specialized orientation extractor: it produces a continuous chromatic shift for the circles (Fig.~\ref{fig:synthetic}b) while maintaining a nearly uniform hue for the horizontal ridges (Fig.~\ref{fig:synthetic}f), as expected for a constant orientation field. Conversely, the \textit{Refinement-Enc-Dec} captures frequency variations: it exhibits a vertical gradient for the varying-frequency pattern (Fig.~\ref{fig:synthetic}g), whereas it yields a homogeneous response for the constant-frequency circles (Fig.~\ref{fig:synthetic}c). The final representation (Figs.~\ref{fig:synthetic}d and~\ref{fig:synthetic}h) synthesizes both attributes, suggesting that LEADER has developed universal signal-processing primitives for quasi-periodic patterns.

\section{Conclusion}
\label{sec:conclusion}
In this work, we introduced LEADER, a lightweight end-to-end architecture for minutiae extraction that maps raw fingerprint images to comprehensive minutiae attributes through an integrated on-graph pipeline with deterministic readout, achieving SOTA performance with only 0.9\,M parameters. Across three heterogeneous benchmarks, the proposed model demonstrates superior precision and reliability, maintaining high $F_1$-scores even under stringent threshold levels, where competing methods degrade significantly. Sample-level ranking analysis confirms this systematic superiority, with LEADER consistently outperforming both deep learning competitors and established commercial systems on individual samples.

A key finding is the model's robust cross-domain generalization. Trained exclusively on plain fingerprints, LEADER learns fundamental topological structures that enable it to outperform specialized systems on both NIST SD27 and SD302 latent datasets. Qualitative visual analysis suggests that internal representations align with core biometric primitives, such as segmentation masks, orientation fields, ridge frequencies, and skeletons. The emergence of these representations, combined with the observed ability to perform topological inpainting over deep ridge cuts, suggests that LEADER captures meaningful structural characteristics of fingerprint morphology beyond local pattern matching. These qualitative observations support the hypothesis that a compact model, guided by the proposed CMR topological supervision, can achieve greater abstraction and generalization than larger, unconstrained architectures.

The model's computational efficiency confirms that high-precision topological reasoning does not require resource-intensive hardware. By combining high inference throughput with a 0.9\,M parameter count and a small memory footprint (less than 20\,MB for intermediate tensors and weights), LEADER is well-suited for local identification on resource-constrained platforms, though physical hardware-in-the-loop embedded deployment remains an objective for future work.

Despite these advances, accurate feature extraction from severely degraded latent evidence remains an open frontier. Sample-level visual analysis (Figs.~\ref{fig:visual_comparison_main}j and~\ref{fig:visual_comparison_main}p) reveals that extreme environmental noise, severe structural fragmentation, and heavy ridge loss can still exceed the reconstructive capabilities of supervised models. Furthermore, LEADER operates at a standard 500\,DPI resolution; while higher-resolution inputs (such as 1000\,DPI latent images from NIST SD302) are effectively handled via resampling, native multi-scale architectural support for varying image resolution remains a limitation. Moreover, while our evaluation framework relies on ground-truth-defined cropping to decouple extraction performance from upstream segmentation biases, assessing performance on fully uncropped inputs or integrated automatic segmentation pipelines constitutes an important avenue for future work. Finally, upcoming investigations will explore self-supervised pre-training on large-scale unlabeled datasets and Transformer-based autoencoders (e.g., masked reconstruction strategies~\cite{Jia2025}) to infer missing ridge structures in highly corrupted impressions, as well as native multi-resolution processing capabilities.

\section*{Acknowledgment}
The authors are grateful to Dr.~Davide Maltoni for his valuable suggestions and insightful discussions.

\bibliographystyle{IEEEtran}
\bibliography{fingerprint}

\vfill
\clearpage

\newpage

\twocolumn[
    \setcounter{page}{1}
    \begin{center}
        \small \textsc{Supplementary Material for} \\
        \vspace{0.5em}
        \Large \textbf{LEADER: Lightweight End-to-End Attention-Gated\\ Dual Encoder-Decoder Network for Robust Minutiae Extraction} \\
        \vspace{0.5em}
        \small Raffaele Cappelli and Matteo Ferrara \\
    \end{center}
    
    \vspace{1em}
]

\appendices

\section{Architecture and Computational Layers}
\label{sec:supp_architecture}

\begin{figure}[bht]
    \centering
    \includegraphics[width=0.85\linewidth]{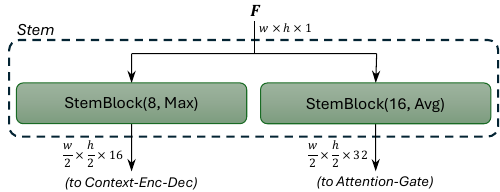}
    \caption{\textbf{\textit{Stem} component.} Detailed view of the dual-path input stage. It accepts the raw fingerprint $\mathbf{F}$ and processes it through two specialized \textit{StemBlocks} (detailed in Fig.~\ref{fig:supp_blocks}d).}
    \label{fig:supp_stem}
\end{figure}

\begin{figure}[ht]
    \centering
    \includegraphics[width=0.85\linewidth]{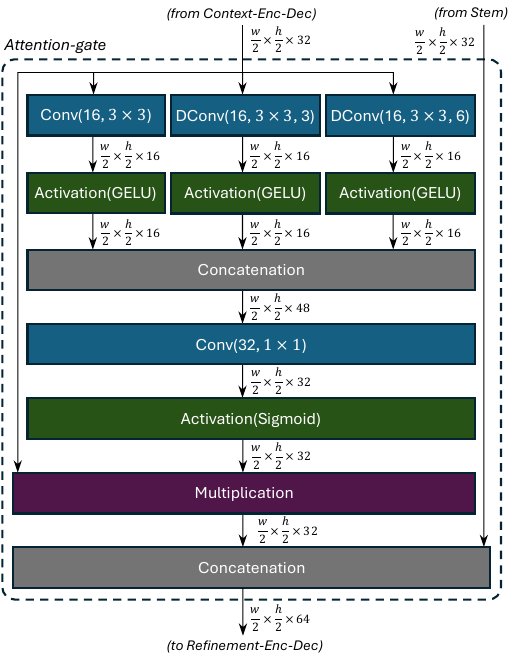}
    \caption{\textbf{\textit{Attention-Gate} component.} Internal flow of the recalibration mechanism. It computes the multi-scale gating signal $\mathbf{X}_{gate}$ through parallel dilated paths and modulates the input features using element-wise multiplication before the final concatenation.}
    \label{fig:supp_ga}
\end{figure}

\begin{figure}[ht]
    \centering
    \includegraphics[width=0.9\linewidth]{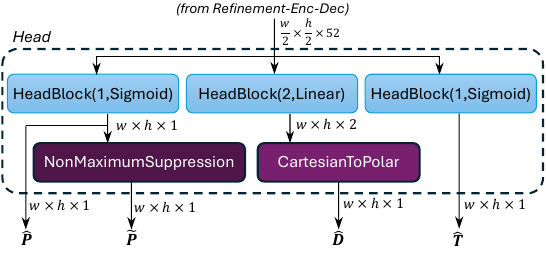}
    \caption{\textbf{\textit{Head} component.} Diagram of the task-specific decoding logic. The shared features are branched into three \textit{HeadBlocks} (detailed in Fig.~\ref{fig:supp_blocks}e) for position, direction, and type estimation. Task-specific activations are applied within each block: 
    Sigmoid for position and type maps and Linear (identity) for the regression of Cartesian directional components. The pipeline concludes with on-graph \textit{NMS} and \textit{CartesianToPolar} angular decoding.}
    \label{fig:supp_head}
\end{figure}

\begin{figure}[ht]
    \centering
    \includegraphics[width=\linewidth]{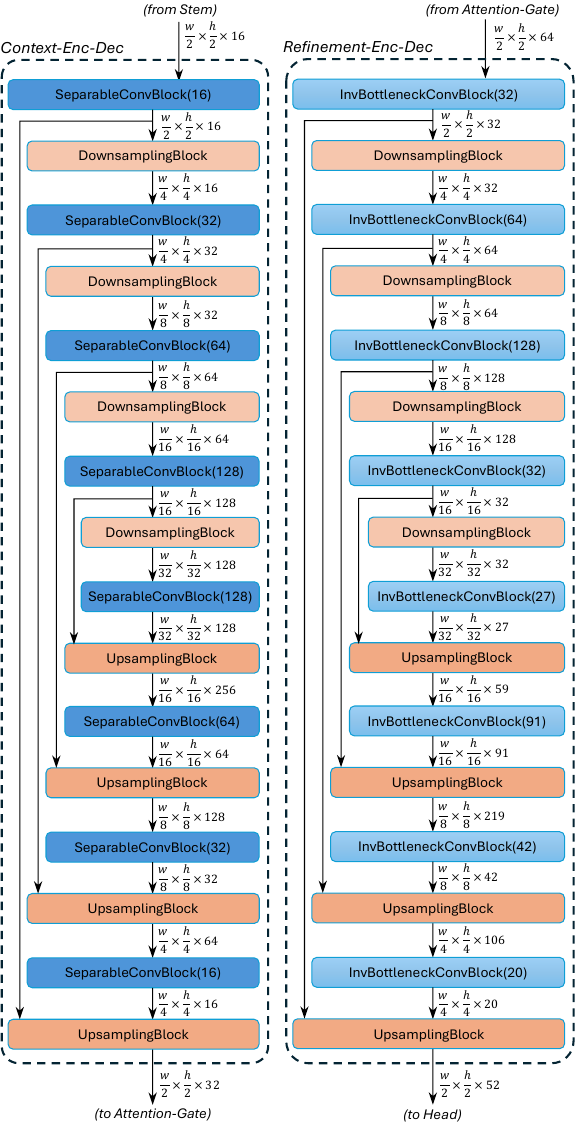}
    \caption{\textbf{\textit{Context-Enc-Dec} and \textit{Refinement-Enc-Dec} components.} Side-by-side comparison: The \textit{Context-Enc-Dec} (left) employs \textit{SeparableConvBlocks} for topological extraction (Fig.~\ref{fig:supp_blocks}c), while the \textit{Refinement-Enc-Dec} (right) uses \textit{InvBottleneckConvBlocks} for high-level semantic refinement (Fig.~\ref{fig:supp_blocks}f). Both stages adopt \textit{DownsamplingBlocks} (Fig.~\ref{fig:supp_blocks}a) and \textit{UpsamplingBlocks} (Fig.~\ref{fig:supp_blocks}b).}
    \label{fig:supp_sae}
\end{figure}

\begin{figure}[t]
    \centering
    \includegraphics[width=0.97\linewidth]{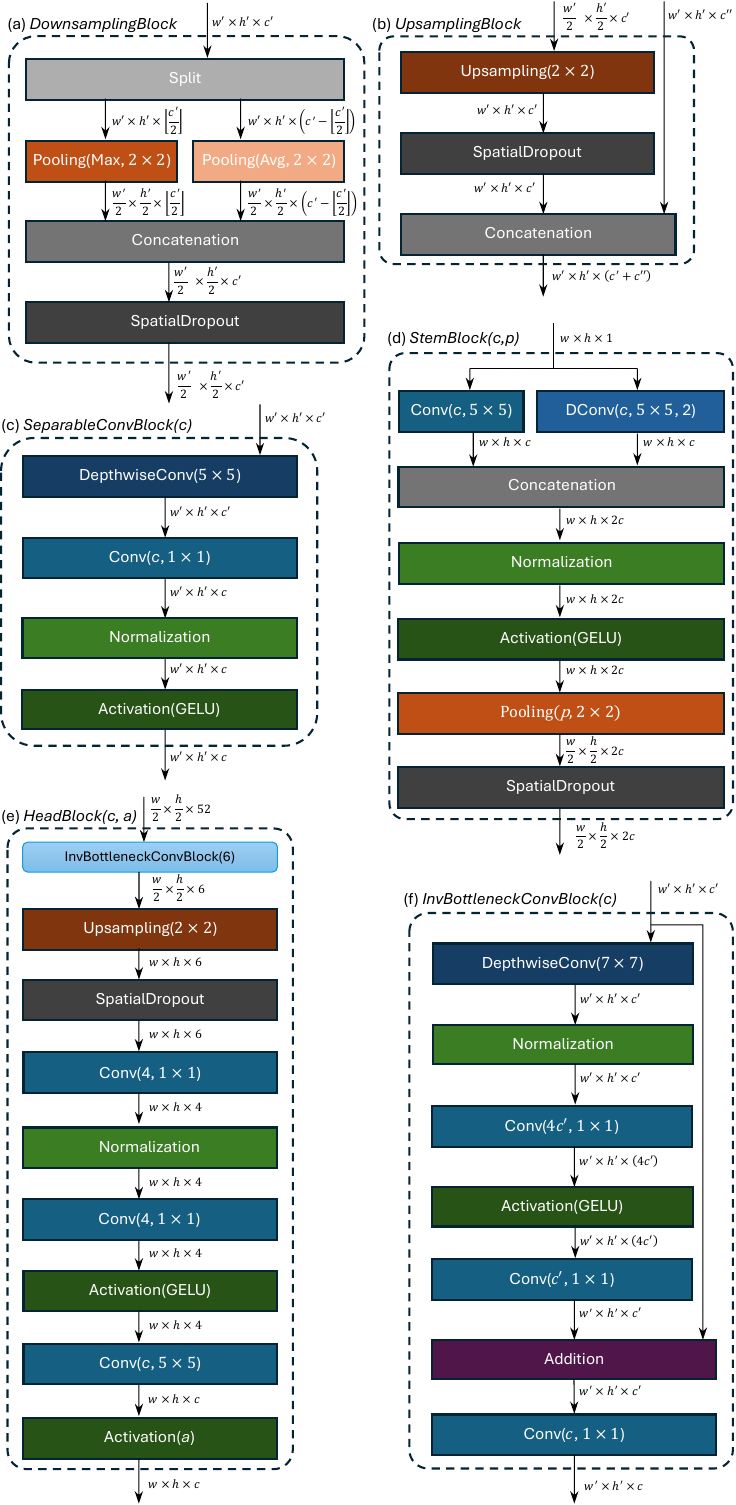}
    \caption{\textbf{Computational block library.} Schematic of the modular blocks in LEADER: (a)~\textit{DownsamplingBlock} and (b)~\textit{UpsamplingBlock} for resolution management; (c)~\textit{SeparableConvBlock} for efficient feature extraction; (d)~\textit{StemBlock} for initial feature extraction and spatial mapping; (e)~\textit{HeadBlock} for task-specific projection; and (f)~\textit{InvBottleneckConvBlock}, the core building block of the \textit{Refinement-Enc-Dec}, designed for high-capacity representation with minimal parameter overhead. For blocks employed at multiple stages, input dimensions are denoted as $w' \times h' \times c'$ to indicate that the operation is resolution-agnostic and applicable to feature maps with varying spatial sizes and channel depths throughout the network.}
    \label{fig:supp_blocks}
\end{figure}

This appendix provides a detailed hierarchical decomposition of the LEADER architecture, as illustrated in Figs.~\ref{fig:supp_stem}--\ref{fig:supp_blocks}, along with the definitions of the primitive computational layers used throughout the network. 

To ensure stable gradient flow and superior non-linear representation compared to traditional ReLU-based pipelines, LEADER consistently employs Layer Normalization~\cite{Ba2016} and GELU activation~\cite{Hendrycks2023}. The following primitive operations form the building blocks of the architecture:

\begin{itemize}[leftmargin=*]
    \small
    \item \textit{Conv}($c, s \times s$): Standard 2D convolution with $c$ filters of size $s \times s$, and bias.
    \item \textit{DConv}($c, s \times s, r$): Dilated 2D convolution with $c$ filters of size $s \times s$, dilation rate $r$, and bias.
    \item \textit{DepthwiseConv}($s \times s$): Parameter-efficient per-channel 2D convolution with kernel size $s \times s$, without bias.
    \item \textit{Normalization}: Layer Normalization applied channel-wise.
    \item \textit{Activation}($a$): Computational layer applying function $a$, which can represent a non-linear activation (typically GELU or Sigmoid) or the identity mapping (Linear).
    \item \textit{Pooling}($p, s \times s$): Spatial downsampling using strategy $p$ (Avg or Max), with an isotropic kernel size and stride of $s$.
    \item \textit{Upsampling}($s \times s$): Spatial upsampling by a factor of $s$ via nearest-neighbor interpolation.
    \item \textit{SpatialDropout}: Regularization layer that randomly masks entire feature maps to prevent co-adaptation.
    \item \textit{Concatenation}\,/\,\textit{Split}: Channel-wise concatenation of feature maps or their separation into two equal halves.
    \item \textit{Multiplication}\,/\,\textit{Addition}: Element-wise operations used for feature gating and residual learning.
    \item \textit{NonMaximumSuppression}: On-graph peak-selection layer (see Sec.~\ref{subsec:postprocessing}).
    \item \textit{CartesianToPolar}: On-graph angular decoding layer (see Sec.~\ref{subsec:postprocessing}).
\end{itemize}

\section{Implementation and Training Details}
\label{sec:training}
LEADER is implemented using the Keras framework~\cite{Chollet2015} with the TensorFlow backend~\cite{TensorFlowDevelopers2015}. To promote a robust learning process across the heterogeneous datasets described in Sec.~\ref{subsec:setup}, we adopt the following protocol.

\subsection{Data Preparation and Augmentation}
To maintain a consistent input size, training samples are obtained by cropping patches of $256 \times 320$ pixels from the source images. Multiple patches per image are extracted to ensure full coverage of the ridge area while minimizing excessive overlap. To improve generalization, an extensive data augmentation pipeline has been applied, including random translations, rotations ($\pm90^\circ$), scaling ($\pm25\%$), and flips, as well as gamma correction and contrast reduction.
In addition, domain-specific perturbations simulate ridge thickness variations (through morphological erosion and dilation), alongside physical artifacts such as scratches and abrasions. During spatial transformations, empty regions are filled using constant padding based on the average background intensity---computed from ground-truth segmentation masks---to prevent high-contrast border artifacts.

\subsection{Optimization and Training Strategy}
LEADER is trained using the Lion optimizer~\cite{Chen2024} for a fixed duration of $25$ epochs, each comprising $400$ batches of size $32$. The momentum parameters ($\beta_1=0.5, \beta_2=0.9$) are calibrated on the tuning set (Table~\ref{tab:datasets}) to ensure stable convergence. The learning rate follows a cosine-decay schedule spanning the range $[10^{-6}, 3 \times 10^{-3}]$, which includes an initial warm-up phase. To mitigate overfitting, we integrate a weight decay of $0.35$ and a spatial dropout rate of $2\%$ (applied to the layers detailed in Appendix~\ref{sec:supp_architecture}). These regularization strategies, combined with the smooth convergence provided by the cosine-decay schedule, allow stable training without the need for a validation-based early stopping criterion. 

Parameters for the CMR ground-truth encoding ($\delta=4, \beta=2, \sigma=2, \lambda=0.3$) and multi-task loss weights ($\alpha_p=0.85, \alpha_d=0.10, \alpha_t=0.05$) are empirically selected using the tuning set (Table~\ref{tab:datasets}). Training takes approximately $40$ minutes on an NVIDIA GeForce RTX™ 3080 Ti GPU.

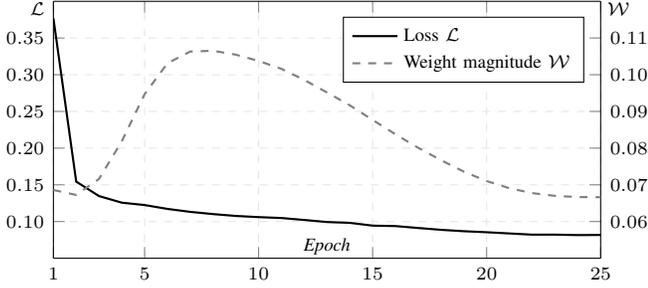
\begin{figure}[htb]
    \centering
    \begin{tikzpicture}
        \pgfplotsset{set layers}
        
        \begin{axis}[
            IEEEstyle,
            height=5cm,
            xmin=1, xmax=25,
            xtick={1,5,10,15,20,25},
            xlabel={\textit{Epoch}},
            xlabel style={at={(0.5,-0.02)}, anchor=south, font=\scriptsize\itshape},
            ymin=0.05, ymax=0.4,
            ytick={0.10, 0.15, 0.20, 0.25, 0.30, 0.35},
            yticklabel style={
                /pgf/number format/fixed,
                /pgf/number format/precision=2,
                /pgf/number format/fixed zerofill
            },            
            ylabel={$\mathcal{L}$},
            ylabel style={at={(-0.06,0.91)}, anchor=south west, rotate=-90, font=\scriptsize},
            axis y line*=left,
            ymajorgrids=true,
            grid style={dashed, gray!20},
            legend cell align={left},
            legend style={
                at={(0.97,0.94)}, 
                anchor=north east, 
                draw=black, 
                fill=white, 
                font=\scriptsize,
                row sep=0pt
            }
        ]
            \addplot[color=black, thick] coordinates {
                (1, 0.3768588602542877) (2, 0.1543891280889511) (3, 0.13469336926937103) (4, 0.12569978833198547) (5, 0.12246295064687729) (6, 0.11721663177013397) (7, 0.1131817027926445) (8, 0.11016172170639038) (9, 0.1076212227344513) (10, 0.10602083802223206) (11, 0.10479236394166946) (12, 0.10212209075689316) (13, 0.09932731837034225) (14, 0.0981951430439949) (15, 0.09442158043384552) (16, 0.09381657838821411) (17, 0.0912066102027893) (18, 0.08870731294155121) (19, 0.08680281043052673) (20, 0.08542989939451218) (21, 0.08376147598028183) (22, 0.08213227987289429) (23, 0.08215703815221786) (24, 0.0815434381365776) (25, 0.08171848952770233)
            };
            \addlegendentry{Loss $\mathcal{L}$}

            \addlegendimage{color=gray, dashed, thick}
            \addlegendentry{Weight magnitude $\mathcal{W}$}
        \end{axis}

        \begin{axis}[
            IEEEstyle,
            height=5cm,
            xmin=1, xmax=25,
            axis x line=none,
            ymin=0.054, ymax=0.124,
            ytick={0.064, 0.074, 0.084, 0.094, 0.104, 0.114},
            yticklabel style={
                /pgf/number format/fixed,
                /pgf/number format/precision=2,
                /pgf/number format/fixed zerofill
            },
            ylabel={$\mathcal{W}$},
            ylabel style={at={(1.07,0.91)}, anchor=south east, rotate=-90, font=\scriptsize},
            axis y line*=right,
            grid=none,
        ]
            \addplot[color=gray, dashed, thick] coordinates {
                (1, 0.072593) (2, 0.071249) (3, 0.075620) (4, 0.085879) (5, 0.098705) (6, 0.107109) (7, 0.110316) (8, 0.110525) (9, 0.109460) (10, 0.107760) (11, 0.105600) (12, 0.102581) (13, 0.099159) (14, 0.095664) (15, 0.091678) (16, 0.087772) (17, 0.084077) (18, 0.080717) (19, 0.077588) (20, 0.075033) (21, 0.073069) (22, 0.071739) (23, 0.070991) (24, 0.070671) (25, 0.070625)
            };
        \end{axis}
    \end{tikzpicture}
    \caption{\textbf{Optimization dynamics during LEADER training.} The primary axis (left) shows the minimization of the multi-task loss function $\mathcal{L}$. The secondary axis (right) illustrates the evolution of the normalized weight magnitude $\mathcal{W}$, indicating stable convergence and controlled model capacity.}
    \label{fig:training_dynamics}
\end{figure}

The optimization dynamics are illustrated in Fig.~\ref{fig:training_dynamics}, which reports the multi-task loss $\mathcal{L}$ and the normalized weight magnitude $\mathcal{W}$ (calculated as the RMS of all trainable parameters) across the training duration. The loss curve exhibits a rapid initial decrease followed by smooth asymptotic convergence, confirming both the efficacy of the selected hyperparameters and the stability of the Lion optimizer in this multi-task setting. Simultaneously, the trajectory of $\mathcal{W}$ indicates a controlled evolution of the model capacity, where the weight magnitude stabilizes after the initial exploration phase.

\begin{notablediff*}[htb]
\centering
\begin{threeparttable}
\caption{Comparative performance ($F_1$-score) across the three threshold levels under type-agnostic and type-aware regimes without boundary filtering (margin and mask). Values in parentheses denote differences relative to Table~\ref{tab:results_f1}.}
\label{tab:results_border_sens_analysis}
\begin{tabular}{l l ccc c ccc c ccc} 
\toprule
& & \multicolumn{3}{c}{\textbf{FVC2002 DB1-A}} & & \multicolumn{3}{c}{\textbf{NIST SD27}} & & \multicolumn{3}{c}{\textbf{NIST SD302}} \\
\cmidrule(lr){3-5} \cmidrule(lr){7-9}  \cmidrule(lr){11-13}
& \textbf{Method} & \thrl{16}{6} & \thrl{12}{8} & \thrl{8}{10} &
                  & \thrl{16}{6} & \thrl{12}{8} & \thrl{8}{10} &
                  & \thrl{16}{6} & \thrl{12}{8} & \thrl{8}{10} \\
\midrule
\multirow{10}{*}{\rotatebox[origin=c]{90}{\textit{Type-agnostic}}} 

& FingerNet & \underline{0.87}  \tiny (-0.00) & \underline{0.86}  \tiny (-0.00) & \underline{0.83}  \tiny (-0.00) & & \underline{0.65}  \tiny (-0.02) & \underline{0.63}  \tiny (-0.01) & \underline{0.56}  \tiny (-0.01) & & \underline{0.64}  \tiny (-0.01) & \underline{0.62}  \tiny (-0.02) & \underline{0.59}  \tiny (-0.01) \\
& LatentAFIS & 0.83 \tiny (-0.01) & 0.83 \tiny (-0.00) & 0.81 \tiny (-0.00) & & 0.61 \tiny (-0.01) & 0.58 \tiny (-0.01) & 0.53 \tiny (-0.01) & & 0.55 \tiny (-0.02) & 0.54 \tiny (-0.02) & 0.50 \tiny (-0.02) \\
& MinutiaeNet & 0.46 \tiny (-0.01) & 0.42 \tiny (-0.01) & 0.37 \tiny (-0.01) & & 0.47 \tiny (-0.01) & 0.44 \tiny (-0.01) & 0.40 \tiny (-0.00) & & 0.24 \tiny (-0.01) & 0.20 \tiny (-0.01) & 0.17 \tiny (-0.00) \\
\addlinespace[0.25em]
& FingerJet & 0.82 \tiny (-0.01) & 0.79 \tiny (-0.01) & 0.74 \tiny (-0.01) & & 0.37 \tiny (-0.03) & 0.34 \tiny (-0.02) & 0.27 \tiny (-0.02) & & 0.21 \tiny (-0.01)& 0.20 \tiny (-0.01) & 0.17 \tiny (-0.01) \\
& FDx & 0.85 \tiny (-0.01) & 0.84 \tiny (-0.01) & 0.80 \tiny (-0.01) & & 0.36 \tiny (-0.09) & 0.33 \tiny (-0.08) & 0.28 \tiny (-0.06) & & 0.35 \tiny (-0.04) & 0.33 \tiny (-0.03) & 0.29 \tiny (-0.04) \\
& VeriFinger & 0.85 \tiny (-0.01) & 0.84 \tiny (-0.01) & 0.79 \tiny (-0.01) & & 0.60 \tiny (-0.04) & 0.57 \tiny (-0.04) & 0.50 \tiny (-0.03) & & 0.56 \tiny (-0.04) & 0.54 \tiny (-0.03) & 0.49 \tiny (-0.03) \\
\addlinespace[0.25em]
& MINDTCT & 0.82 \tiny (-0.02) & 0.80 \tiny (-0.02) & 0.76 \tiny (-0.02) & & 0.31 \tiny (-0.01) & 0.28 \tiny (-0.01) & 0.23 \tiny (-0.01) & & 0.29 \tiny (-0.01) &0.26 \tiny (-0.02) & 0.23 \tiny (-0.02) \\
& FingerFlow & 0.53 \tiny (-0.00) & 0.49 \tiny (-0.00) & 0.42 \tiny (-0.01) & & 0.49 \tiny (-0.00) & 0.46 \tiny (-0.01) & 0.42 \tiny (-0.00) & & 0.25 \tiny (-0.01) & 0.21 \tiny (-0.01) & 0.18 \tiny (-0.00) \\
& SourceAFIS & 0.75 \tiny (-0.03) & 0.74 \tiny (-0.02) & 0.70 \tiny (-0.02) & & 0.23 \tiny (-0.06) & 0.20 \tiny (-0.05) & 0.15 \tiny (-0.04) & & 0.25 \tiny (-0.03) & 0.22 \tiny (-0.02) & 0.18 \tiny (-0.02) \\
\addlinespace[0.25em]
& LEADER (ours) & \textbf{0.91}  \tiny (-0.01) & \textbf{0.90}  \tiny (-0.01) & \textbf{0.90}  \tiny (-0.00) & & \textbf{0.69}  \tiny (-0.02) & \textbf{0.67}  \tiny (-0.02) & \textbf{0.61}  \tiny (-0.01) & & \textbf{0.65}  \tiny (-0.04) & \textbf{0.64}  \tiny (-0.03) & \textbf{0.62}  \tiny (-0.02) \\
\midrule
\multirow{7}{*}{\rotatebox[origin=c]{90}{\textit{Type-aware}}}
& FingerJet & 0.64 \tiny (-0.00) & 0.61 \tiny (-0.01) & 0.58 \tiny (-0.00) & & 0.25 \tiny (-0.02) & 0.23 \tiny (-0.02) & 0.19 \tiny (-0.01) & & 0.15 \tiny (-0.01)& 0.14 \tiny (-0.01) & 0.13 \tiny (-0.00) \\
& FDx & \underline{0.72}  \tiny (-0.01) & \underline{0.71}  \tiny (-0.00) & \underline{0.68}  \tiny (-0.01) & & 0.24 \tiny (-0.06) & 0.22 \tiny (-0.06) & 0.19 \tiny (-0.05) & & 0.24 \tiny (-0.03) & 0.23 \tiny (-0.02) & 0.21 \tiny (-0.02) \\
& VeriFinger & 0.70 \tiny (-0.01) & 0.69 \tiny (-0.01) & 0.66 \tiny (-0.01) & & \underline{0.39}  \tiny (-0.03) & \underline{0.37}  \tiny (-0.02) & \underline{0.33}  \tiny (-0.02) & & \underline{0.38}  \tiny (-0.03) & \underline{0.36}  \tiny (-0.03) & \underline{0.33}  \tiny (-0.03) \\
\addlinespace[0.25em]
& MINDTCT & 0.70 \tiny (-0.02) & 0.68 \tiny (-0.02) & 0.65 \tiny (-0.02) & & 0.20 \tiny (-0.01) & 0.19 \tiny (-0.00) & 0.16 \tiny (-0.01) & & 0.21 \tiny (-0.01) &0.19 \tiny (-0.01) & 0.17 \tiny (-0.01) \\
& FingerFlow & 0.34 \tiny (-0.00) & 0.31 \tiny (-0.00) & 0.27 \tiny (-0.00) & & 0.29 \tiny (-0.00) & 0.27 \tiny (-0.00) & 0.25 \tiny (-0.00) & & 0.17 \tiny (-0.00) & 0.14 \tiny (-0.00) & 0.12 \tiny (-0.00) \\
& SourceAFIS & 0.60 \tiny (-0.02) & 0.58 \tiny (-0.02) & 0.56 \tiny (-0.02) & & 0.15 \tiny (-0.04) & 0.13 \tiny (-0.04) & 0.10 \tiny (-0.03) & & 0.16 \tiny (-0.02) & 0.14 \tiny (-0.02) & 0.12 \tiny (-0.01) \\
\addlinespace[0.25em]
& LEADER (ours) & \textbf{0.84}  \tiny (-0.01) & \textbf{0.84}  \tiny (-0.00) & \textbf{0.83}  \tiny (-0.00) & & \textbf{0.51}  \tiny (-0.01) & \textbf{0.49}  \tiny (-0.02) & \textbf{0.46}  \tiny (-0.01) & & \textbf{0.50}  \tiny (-0.03) & \textbf{0.49}  \tiny (-0.03) & \textbf{0.48}  \tiny (-0.02) \\

\bottomrule
\end{tabular}
\begin{tablenotes}
\item \textbf{Bold} and \underline{underlined} values indicate best and second-best scores, respectively.
\end{tablenotes}
\end{threeparttable}
\end{notablediff*}

\section{Boundary-Filtering Sensitivity Analysis}
\label{sec:border_sens_analysis}

To evaluate the robustness of the comparative findings against boundary artifacts, we conducted an additional sensitivity analysis by completely disabling boundary filtering (i.e., removing both the 14\,px boundary exclusion margin and the filtering of minutiae falling outside the ground-truth segmentation mask). Note that input images remain cropped to the ridge-area bounding box defined by ground-truth masks. As summarized in Table~\ref{tab:results_border_sens_analysis}, LEADER's relative performance rankings remain entirely stable across all datasets.

\section{Extended Performance Analysis}
\label{sec:supp_results}
\begin{figure*}[t]
    \centering
    \begin{minipage}{\textwidth}
        \centering
        \includegraphics[width=\textwidth]{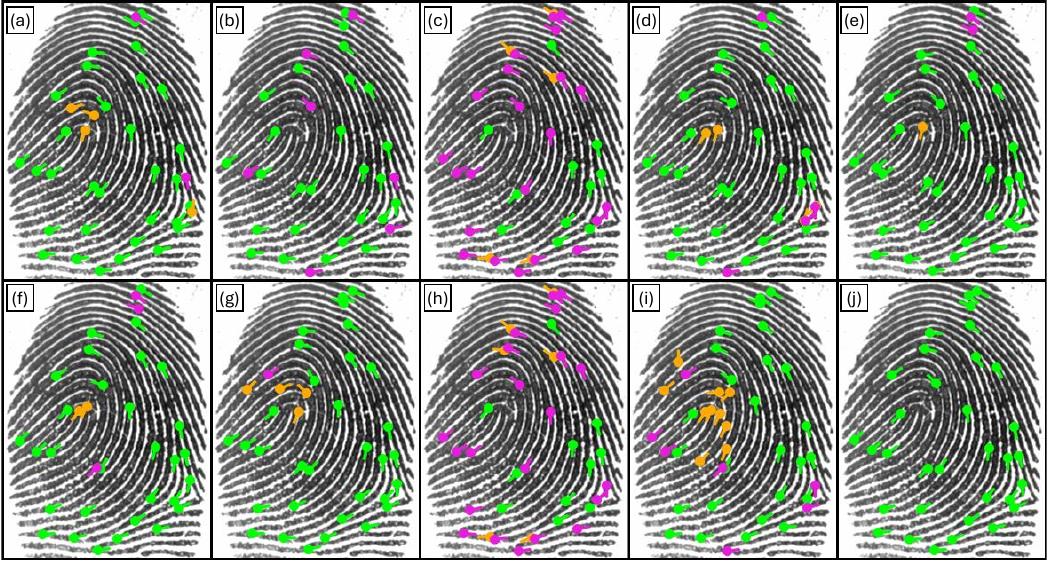}
        \caption{\textbf{Visual comparison on an FVC2002 DB1-A sample representative of high-performance results for the proposed method.} True Positives (green), False Positives (orange), and False Negatives (magenta) are overlaid for each extractor. The metrics (TP\,/\,FP\,/\,FN\,/\,$F_1$-score) provide a sample-level performance breakdown: 
    (a)~FingerNet: 31\,/\,4\,/\,2\,/\,0.91; 
    (b)~LatentAFIS: 26\,/\,0\,/\,7\,/\,0.88; 
    (c)~MinutiaeNet: 11\,/\,5\,/\,22\,/\,0.45; 
    (d)~FingerJet: 29\,/\,4\,/\,4\,/\,0.88; 
    (e)~FDx: 31\,/\,1\,/\,2\,/\,0.95; 
    (f)~VeriFinger: 30\,/\,2\,/\,3\,/\,0.92; 
    (g)~MINDTCT: 32\,/\,4\,/\,1\,/\,0.93; 
    (h)~FingerFlow: 11\,/\,6\,/\,22\,/\,0.44; 
    (i)~SourceAFIS: 26\,/\,10\,/\,7\,/\,0.75; 
    (j)~LEADER (Ours): 33\,/\,0\,/\,0\,/\,1.0.}
    \label{fig:good_res_plain}
    \end{minipage}

    \vspace{0.5cm}

    \begin{minipage}{\textwidth}
        \centering
        \includegraphics[width=\textwidth]{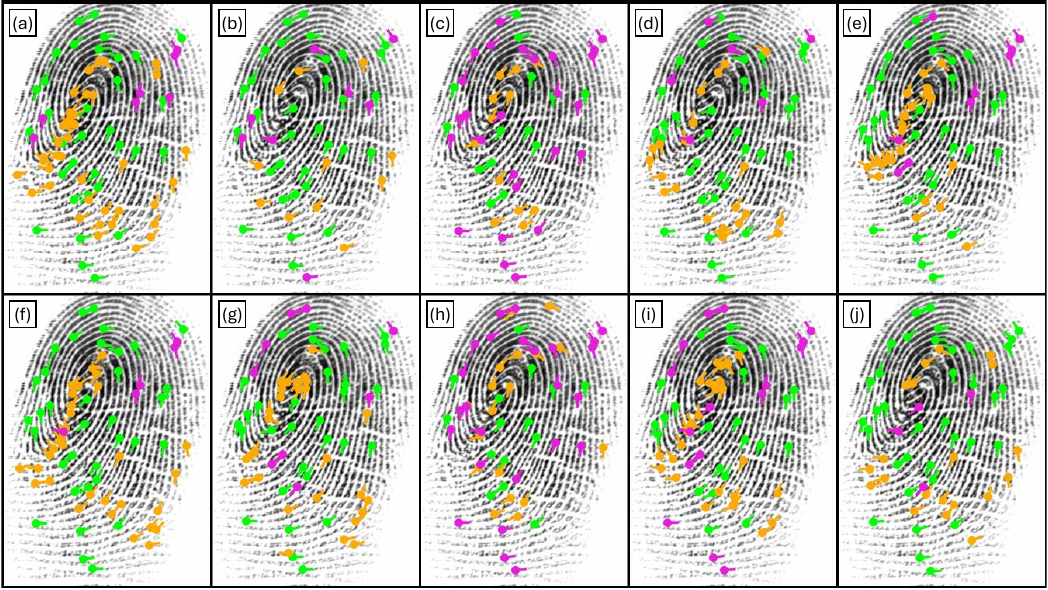}
        \caption{\textbf{Visual comparison on an FVC2002 DB1-A sample representative of low-performance results for the proposed method.} Visualization logic and metric format follow Fig.~\ref{fig:good_res_plain}. 
    (a)~FingerNet: 33\,/\,39\,/\,7\,/\,0.59; 
    (b)~LatentAFIS: 32\,/\,11\,/\,8\,/\,0.77; 
    (c)~MinutiaeNet: 14\,/\,12\,/\,26\,/\,0.42; 
    (d)~FingerJet: 34\,/\,22\,/\,6\,/\,0.71; 
    (e)~FDx: 31\,/\,24\,/\,9\,/\,0.65; 
    (f)~VeriFinger: 35\,/\,32\,/\,5\,/\,0.65; 
    (g)~MINDTCT: 31\,/\,28\,/\,9\,/\,0.63; 
    (h)~FingerFlow: 15\,/\,18\,/\,25\,/\,0.41; 
    (i)~SourceAFIS: 24\,/\,28\,/\,16\,/\,0.52; 
    (j)~LEADER (Ours): 36\,/\,20\,/\,4\,/\,0.75.
    Note that several False Positives in this specific case are attributable to omissions in the ground truth rather than extraction failures.}
        \label{fig:bad_res_plain}
    \end{minipage}
\end{figure*}

\begin{figure*}[t]
    \centering
    \begin{minipage}{\textwidth}
        \centering
        \includegraphics[width=\textwidth]{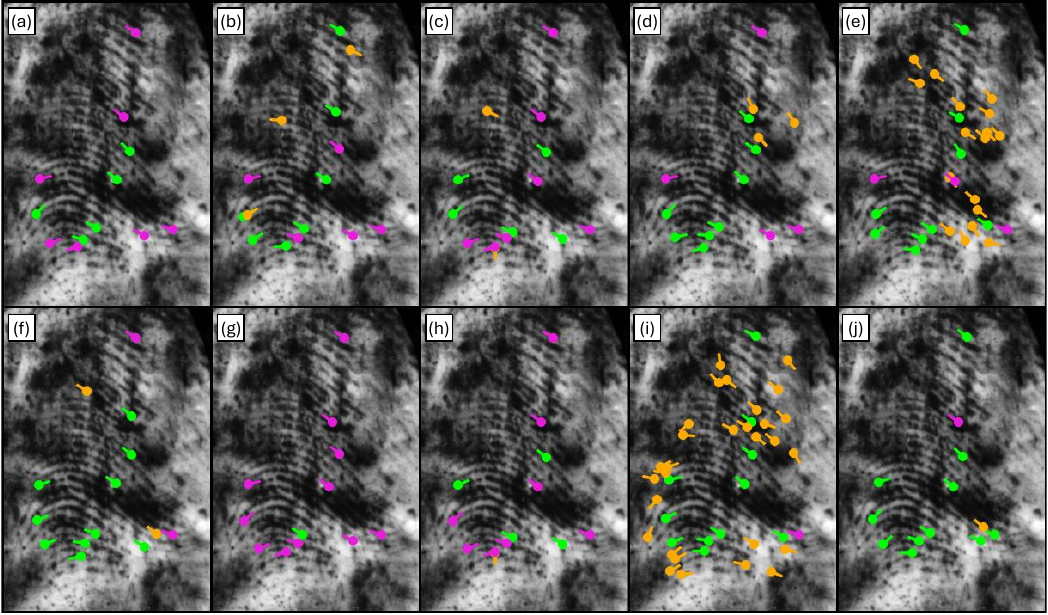}
        \caption{\textbf{Visual comparison on a NIST SD27 latent sample representative of high-performance results for the proposed method.} Visualization logic and metric format follow Fig.~\ref{fig:good_res_plain}. 
    (a)~FingerNet: 5\,/\,0\,/\,7\,/\,0.59; 
    (b)~LatentAFIS: 7\,/\,3\,/\,5\,/\,0.64; 
    (c)~MinutiaeNet: 5\,/\,2\,/\,7\,/\,0.53; 
    (d)~FingerJet: 8\,/\,3\,/\,4\,/\,0.70; 
    (e)~FDx: 9\,/\,17\,/\,3\,/\,0.47; 
    (f)~VeriFinger: 10\,/\,2\,/\,2\,/\,0.83; 
    (g)~MINDTCT: 1\,/\,0\,/\,11\,/\,0.15; 
    (h)~FingerFlow: 4\,/\,1\,/\,8\,/\,0.47; 
    (i)~SourceAFIS: 11\,/\,29\,/\,1\,/\,0.42; 
    (j)~LEADER (Ours): 11\,/\,1\,/\,1\,/\,0.92.}
        \label{fig:good_res_latent}
    \end{minipage}

    \vspace{1cm}

    \begin{minipage}{\textwidth}
        \centering
        \includegraphics[width=\textwidth]{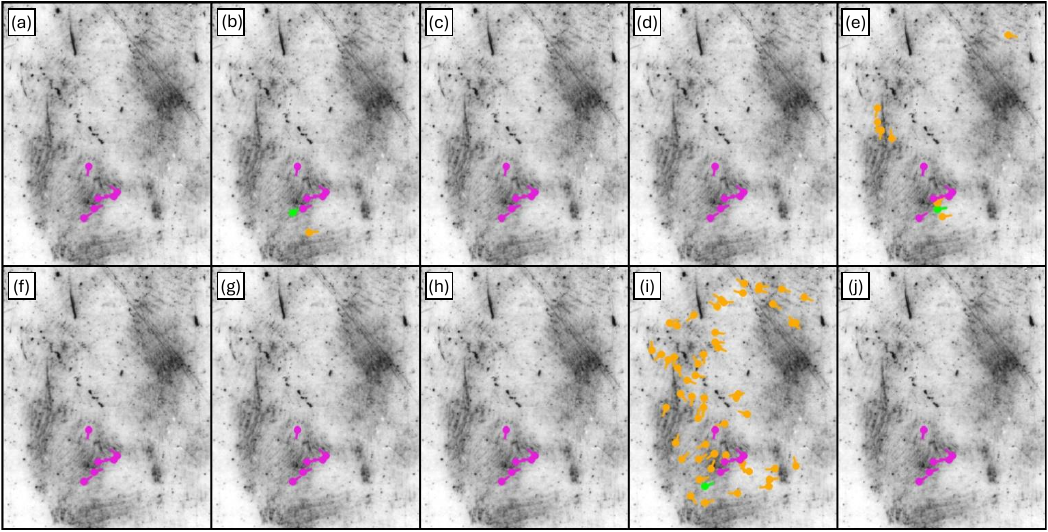}
        \caption{\textbf{Visual comparison on a NIST SD27 latent sample representative of low-performance results for the proposed method.} Visualization logic and metric format follow Fig.~\ref{fig:good_res_plain}. 
    (a)~FingerNet: 0\,/\,0\,/\,6\,/\,0.00; 
    (b)~LatentAFIS: 1\,/\,1\,/\,5\,/\,0.25; 
    (c)~MinutiaeNet: 0\,/\,0\,/\,6\,/\,0.00; 
    (d)~FingerJet: 0\,/\,0\,/\,6\,/\,0.00; 
    (e)~FDx: 1\,/\,7\,/\,5\,/\,0.14; 
    (f)~VeriFinger: 0\,/\,0\,/\,6\,/\,0.00; 
    (g)~MINDTCT: 0\,/\,0\,/\,6\,/\,0.00; 
    (h)~FingerFlow: 0\,/\,0\,/\,6\,/\,0.00; 
    (i)~SourceAFIS: 1\,/\,54\,/\,5\,/\,0.03; 
    (j)~LEADER (Ours): 0\,/\,0\,/\,6\,/\,0.00.}
        \label{fig:bad_res_latent}
    \end{minipage}
\end{figure*}

\begin{figure*}[t]
    \centering
    \begin{minipage}{\textwidth}
        \centering
        \includegraphics[width=\textwidth]{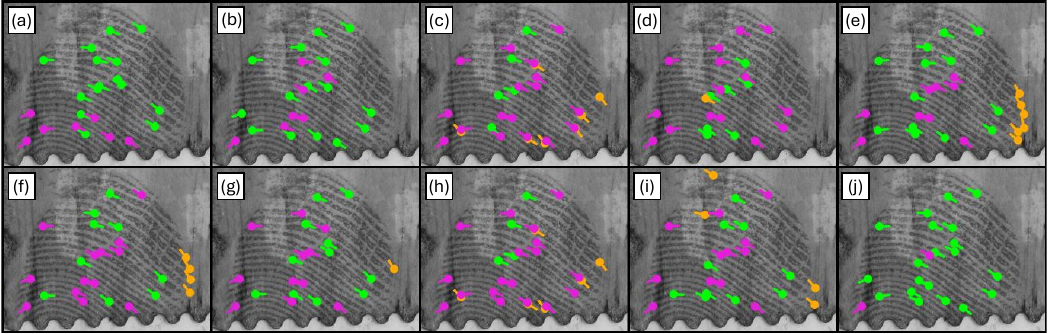}
        \caption{\textbf{Visual comparison on a NIST SD302 latent sample representative of high-performance results for the proposed method.} Visualization logic and metric format follow Fig.~\ref{fig:good_res_plain}. (a)~FingerNet: 15\,/\,0\,/\,7\,/\,0.81; (b)~LatentAFIS: 15\,/\,0\,/\,7\,/\,0.81; (c)~MinutiaeNet: 5\,/\,7\,/\,17\,/\,0.29; (d)~FingerJet: 7\,/\,1\,/\,15\,/\,0.47; (e)~FDx: 13\,/\,6\,/\,9\,/\,0.63; (f)~VeriFinger: 9\,/\,4\,/\,13\,/\,0.51; (g)~MINDTCT: 11\,/\,1\,/\,11\,/\,0.65; (h)~FingerFlow: 4\,/\,6\,/\,18\,/\,0.25; (i)~SourceAFIS: 10\,/\,4\,/\,12\,/\,0.56; (j)~LEADER (ours): 20\,/\,0\,/\,2\,/\,0.95.}
        \label{fig:good_res_latent_302}
    \end{minipage}

    \vspace{0.5cm}

    \begin{minipage}{\textwidth}
        \centering
        \includegraphics[width=\textwidth]{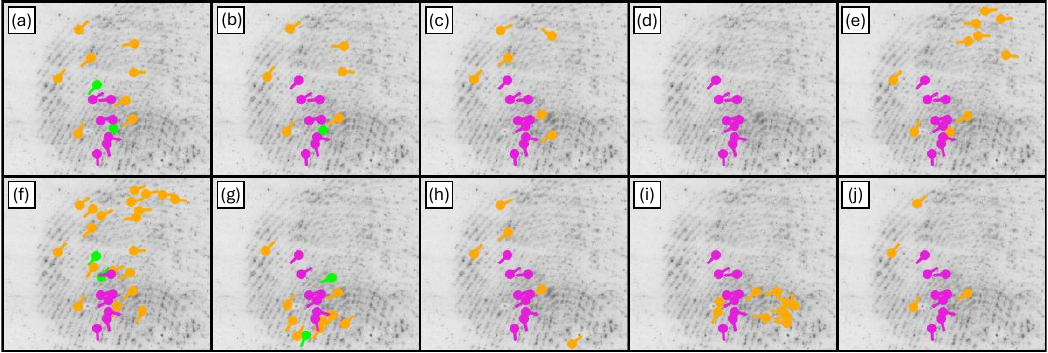}
        \caption{\textbf{Visual comparison on a NIST SD302 latent sample representative of low-performance results for the proposed method.} Visualization logic and metric format follow Fig.~\ref{fig:good_res_plain}. (a)~FingerNet: 2\,/\,8\,/\,7\,/\,0.21; (b)~LatentAFIS: 1\,/\,6\,/\,8\,/\,0.12; (c)~MinutiaeNet: 0\,/\,6\,/\,9\,/\,0.0; (d)~FingerJet: 0\,/\,0\,/\,9\,/\,0.0; (e)~FDx: 0\,/\,10\,/\,9\,/\,0.0; (f)~VeriFinger: 2\,/\,20\,/\,7\,/\,0.13; (g)~MINDTCT: 2\,/\,9\,/\,7\,/\,0.2; (h)~FingerFlow: 0\,/\,5\,/\,9\,/\,0.0; (i)~SourceAFIS: 0\,/\,10\,/\,9\,/\,0.0; (j)~LEADER (ours): 0\,/\,4\,/\,9\,/\,0.0.}
        \label{fig:bad_res_latent_302}
    \end{minipage}
\end{figure*}

To further evaluate the robustness of LEADER against the competitors, this section provides sample-level visual comparisons on the three test databases. 
Figs.~\ref{fig:good_res_plain}--\ref{fig:bad_res_latent_302} provide six sample-level qualitative and quantitative assessments. These examples represent high- and low-performance cases (as determined by the $F_1$-score) and offer a direct comparison of extraction accuracy under varying quality conditions.

\end{document}